\documentclass[sigconf]{acmart}

\usepackage{booktabs}          
\usepackage[table,xcdraw]{xcolor}
\usepackage{tabularx}          
\usepackage{adjustbox}
\usepackage{enumitem}
\usepackage{multirow}

\usepackage{subcaption}

\definecolor{mygreen1}{HTML}{D5E8D4} 
\definecolor{mygreen2}{HTML}{A3D9A5} 
\definecolor{mygreen3}{HTML}{6FBE7A} 
\definecolor{mygreen4}{HTML}{3FA85B} 

\definecolor{myblue1}{HTML}{D9EBF2}  
\definecolor{myblue2}{HTML}{A3C9E6}  
\definecolor{myblue3}{HTML}{6FAAD6}  
\definecolor{myblue4}{HTML}{3F8FC7}  

\definecolor{myred1}{HTML}{F2D9D9}   
\definecolor{myred2}{HTML}{E6A3A3}   
\definecolor{myred3}{HTML}{D67070}   
\definecolor{myred4}{HTML}{C34242}   

\definecolor{lightblue}{RGB}{218, 227, 243}

\AtBeginDocument{%
  }

\begin{document}

\title{Questionnaire meets LLM: A Benchmark and Empirical Study of\\Structural Skills
for Understanding Questions and Responses}

\author{Duc-Hai Nguyen}
\email{125109073@umail.ucc.ie}
\affiliation{%
  \institution{University College Cork}
  \country{Ireland}
}

\author{Vijayakumar Nanjappan}
\email{vnanjappan@ucc.ie}
\affiliation{%
  \institution{University College Cork}
  \country{Ireland}
}

\author{Barry O'Sullivan}
\email{b.osullivan@cs.ucc.ie}
\affiliation{%
  \institution{University College Cork}
  \country{Ireland}
}

\author{Hoang D. Nguyen}
\email{hn@cs.ucc.ie}
\affiliation{%
  \institution{University College Cork}
  \country{Ireland}
}


\begin{abstract}
Millions of people take surveys every day, from market polls and academic studies to medical questionnaires and customer feedback forms. These datasets capture valuable insights, but their scale and structure present a unique challenge for large language models (LLMs), which otherwise excel at few-shot reasoning over open-ended text.
Yet, their ability to process questionnaire data or lists of questions crossed with hundreds of respondent rows remains underexplored.
Current retrieval and survey analysis tools (e.g., Qualtrics, SPSS, REDCap) are typically designed for humans in the workflow, limiting such data integration with LLM and AI-empowered automation.
This gap leaves scientists, surveyors, and everyday users without evidence-based guidance on how to best represent questionnaires for LLM consumption.
We address this by introducing \textbf{QASU} (Questionnaire Analysis and Structural Understanding), a benchmark that probes six structural skills, including answer lookup, respondent count, and multi-hop inference, across six serialization formats and multiple prompt strategies. Experiments on contemporary LLMs show that choosing an effective format and prompt combination can improve accuracy by up to 8.8\% compared to suboptimal formats. For specific tasks, carefully adding a lightweight structural hint through self-augmented prompting can yield further improvements of 3--4\% on average.
By systematically isolating format and prompting effects, our open source\footnote{\url{https://github.com/ReML-AI/QASU}} benchmark offers a simple yet versatile foundation for advancing both research and real-world practice in LLM-based questionnaire analysis.

\end{abstract}




\maketitle
\pagestyle{plain}

\section{INTRODUCTION}

Survey questionnaires collect responses from millions of people on a daily basis, whether it is a market poll, an academic study, a hospital satisfaction form, or a quick in-app feedback request \cite{goodfellow2023overview}. This ongoing flow of answers provides essential insights that shape decisions in many areas. Market researchers use them to understand consumer preferences, healthcare teams track patient satisfaction to improve care quality, and educators rely on them to assess learning outcomes. As survey data accumulates at scale, the challenge is no longer just collecting it, but also making sense of it efficiently and accurately.

From this perspective, structured data can be applied to turn plain text sentences into clear, organized blocks, making it more convenient for algorithms as well as humans to search, group, and analyze. A survey is a familiar example where each row records a single respondent, each column corresponds to a question, and each cell holds the person’s answer. These simple structures power everyday tasks such as counting how many people completed a survey, looking up an answer to a specific question, or finding everyone who gave the same response. Such operations are the foundation of questionnaire analysis, user-experience research, and policy evaluation \cite{groves2009survey,gweon2024bert,rink2024absa}. For scientists, social researchers, healthcare analysts, and even casual poll creators, questionnaires are often the primary evidence base from which decisions are made, making their accessibility and interpretability crucial. Building on this foundation, integrating questionnaires directly with large language models (LLMs) unlocks far more than traditional analysis. An LLM connected to raw survey data could instantly generate executive summaries, craft visually engaging presentations, reason over complex answer patterns, or act as an autonomous agent that can plan and decide the follow-up actions. Envision a hospital using patient satisfaction forms: an LLM could not only aggregate scores but could also point out recurring complaints, summarize them in plain language, and even recommend targeted interventions. In market research, it could infer consumer trends and draft product positioning strategies within minutes. By automating and enriching these workflows, LLMs have the potential to turn static survey data into an active decision-making tool.

Given their importance, it is natural to ask whether emerging AI tools can take over some of the analytical work involved in processing questionnaires. 
Innovations like chain-of-thought reasoning \cite{wei2022cot} and self-consistency sampling \cite{wang2023sc} have since led to even better performance on general benchmarks like MMLU \cite{hendrycks2021mmlu} and table reasoning challenges like TabFact \cite{chen2020tabfact}. However, most prior work has focused on clean, database-like tables \cite{herzig2020tapas, yin2020tabert, chen2021finqa}, which typically arrive preprocessed with uniform schemas and resolved categorical values. Questionnaires, by contrast, retain structural complexities that distinguish them from such benchmarks: mixed answer types (e.g., multiple choice, Likert scales, free text), grouped or nested questions, and skip logic that changes which questions appear for different respondents. While existing table reasoning datasets assume this complexity has been manually cleaned away, QASU evaluates LLM performance on questionnaire data in its raw, pre-processed state, where models must handle heterogeneous data types, resolve answer codes via schema metadata, and interpret questions whose response sets vary across respondents.

These structural differences mean that even before an LLM can attempt any reasoning, the questionnaire must first be prepared in a way the LLM can understand. This requires flattening the data into text, which is a step called \emph{serialization}. This might seem like a minor formatting choice, but in practice, it is a major bottleneck. There is no single standard for how questionnaires should be serialized. Social sciences often use the XML-based DDI Codebook \cite{vardigan2008ddi}; linked-data projects prefer JSON-LD or Turtle triples \cite{scandolari2021surveyontology}; healthcare systems use HL7 FHIR JSON \cite{hl7questionnaire}; and many researchers simply export plain CSV or TSV files \cite{gweon2024bert,rink2024absa}. 
Because every study or organization picks a format independently, it is unclear which serialization style actually helps an LLM best understand and reason over questionnaire data.
This uncertainty has practical consequences since analysts are already experimenting with LLMs to automate tedious questionnaire tasks: extracting answers from free text, checking for inconsistencies, running quick cross-tabulations, or answering ad hoc queries without coding (e.g., asking LLMs to code open-ended responses in surveys \cite{heyde2025surveycoding}; in market research, GPT-3.5 outputs have been used to estimate willingness-to-pay and match them with data from human respondents \cite{brand2024marketresearch}). Yet, existing survey tools such as Qualtrics search, SPSS, or REDCap are built for human use. They perform well at form design and standard reporting, but were not designed to feed structured inputs into LLMs (e.g., Qualtrics may limit real-time API access and require programming knowledge to work around these limitations \cite{behrend2025participant}).

To address these challenges, we introduce \textbf{QASU}: A benchmark that evaluates LLM performance on six micro-tasks, from counting respondents to reverse look-ups, across six serialization styles (HTML, XML, JSON, Markdown, plain text, and Turtle). 
Our results reveal three clear trends:
(1) Format choice significantly impacts performance: HTML achieves up to 8.8\% higher accuracy than TTL on answer lookup tasks, and removing partition marks can degrade performance by 16--24\% across different tasks.
(2) One-shot prompting is critical: zero-shot approaches show 10--25\% lower accuracy, with the gap especially pronounced on multi-hop reasoning tasks (up to 25\% degradation).
(3) Adopting self-augmented prompting \cite{sui2024tablemeetsllm} with lightweight structural hints further boosts performance by 3--4\% on average, particularly benefiting reverse look-ups and conceptual aggregation tasks.  

These findings are not only of academic interest but they also have direct implications for fields like medical research, where accurate processing of clinical questionnaires or patient feedback surveys can inform treatment evaluation, improve hospital services, and guide public health policy. By identifying effective serialization and prompting strategies, our results provide a concrete pathway for researchers and practitioners to integrate LLMs into high-stakes domains with confidence.

\textbf{Our contributions are:}
\begin{itemize}[topsep=0pt]
    \item QASU, the first benchmark to isolate multiple structural skills of LLMs on questionnaire data.
    \item We provide a complete ablation of serialization and prompt options, turning what is now guesswork into practical guidance (see the \S~\ref{sec:experiments}).
    \item Demonstration that self-augmented prompting \cite{sui2024tablemeetsllm} is a simple, model-agnostic method to improve questionnaire reasoning, with benefits for both expert analysts and non-technical users.
\end{itemize}

\section{PRELIMINARIES}

\subsection{Questionnaire Data Structure}

Questionnaire data store answers from multiple respondents to a fixed set of questions, yet the way these questions and answers are arranged can vary greatly across domains, disciplines, and survey platforms. In one of the most common form which is often adopted by large-scale public surveys such as the European Social Survey and the U.S. General Social Survey, the data appear as a flat respondent-question matrix: each row represents one respondent, and each column corresponds to a question or demographic field \cite{groves2009survey}. More complex instruments introduce structural features such as skip logic, multi-select grids, matrix questions, or nested sections \cite{couper2000websurvey, dillman2014internet}, which create conditional flows and hierarchical relationships between items. In this paper, we focus primarily on the flat format, both for its prevalence and its comparability across datasets, while acknowledging that richer, hierarchical formats exist; these are briefly addressed in the \S~\ref{sec:experiments}.

Even within a fixed structural layout, the contents of individual cells vary widely: Plain text is used for open-ended comments, “other” responses, and question wording; numeric values encode Likert-scale ratings, ages, or income ranges; categorical labels mark multiple-choice selections; and date/time fields record events or completion times. Textual elements carry meaning both in the data region (verbatim answers) and in metadata (section headers, instructions), while numeric fields enable statistical operations such as counts, averages, or proportions \cite{fowler2013survey}. The coexistence of heterogeneous data types in which each cell value is tied to the semantics of its column, creates a representational gap between the rigid two-dimensional grid of a questionnaire and the linear token sequence expected by LLMs. Bridging this gap requires preserving both the substantive content and the structural relationships among questions and answers, a challenge also noted in prior work on structured-to-text reasoning \cite{herzig2020tapas, yin2020tabert}.

\subsection{Questionnaire Serialization and Splitting}

Serialization refers to the process of converting a two-dimensional questionnaire into a single linear sequence suitable for LLM input. The most straightforward approach is to write each row in a format in which separates them with a delimiter, like CSV or TSV. However, serialization practices vary greatly across communities and often reflect domain-specific standards. In the social sciences, the DDI Codebook standard wraps questions and answers in XML tags \cite{vardigan2008ddi}; linked-data projects use JSON-LD or Turtle triples to support semantic interoperability \cite{scandolari2021surveyontology}; the healthcare sector employs HL7 FHIR JSON to flatten structured forms into machine-readable resources \cite{hl7questionnaire}. In applied research, open-ended coding pipelines often feed raw TSV directly into text classifiers \cite{gweon2024bert}, whereas aspect-based sentiment analysis studies have inserted Markdown-formatted question blocks directly into prompts for annotation \cite{rink2024absa}. The QASU benchmark (\S~\ref{sec:qasu}) evaluates six such serialization formats under identical conditions, enabling a fair comparison of their impact on LLM performance.

Long-form questionnaires introduce an additional challenge: context length limitations. Transformer-based LLMs have self-attention complexity that scales quadratically with input length \cite{vaswani2017attention}, and even architectures with sparse or linearized attention \cite{beltagy2020longformer, zaheer2020bigbird} face efficiency bottlenecks when encoding thousands of respondents or hundreds of questions in a single pass. Instead of truncating from fixed positions, which risks systematically discarding rare but important responses, we randomly sample a calculated number of respondents for every case that exceeds the token budget. The sample size is determined so that, after serialization, there is guaranteed space to include a one-shot example for in-context guidance, a strategy shown to improve model stability and accuracy in few-shot settings \cite{brown2020gpt3, liu2023gpt4survey}. This approach preserves the original distribution of responses, avoids positional bias, and ensures a consistent prompt structure. Together with our set of serialization baselines, it defines the controlled evaluation environment for QASU.

\section{QASU BENCHMARK}
\label{sec:qasu}
This section introduces a controlled benchmark for systematically examining how input design affects an LLM’s ability to process questionnaire data. We aim to answer two practical questions:  
(1) \emph{Which combination of serialization format, delimiter choice, role label, and format note best enables a model to recognize the question-respondent data format?}  
(2) \emph{Once an effective prompt is established, how well can current models handle the core analytical operations that survey analysts perform daily?}  

The motivation is grounded in real-world practice. Prior studies have shown that GPT-3.5 can code open-ended survey responses at scale \cite{gweon2024bert} and extract aspect sentiment from HR questionnaires \cite{rink2024absa}, yet these works differ substantially in input preparation. Some export data as CSV files, while others format it as Markdown blocks, which makes their results hard to compare directly. QASU removes these inconsistencies by systematically crossing six serialization formats with five prompt configurations and evaluating models on six micro-tasks drawn from routine spreadsheet and survey-analysis workflows. Details on dataset collection and preprocessing appear in \S~\ref{datacollect}.

\subsection{Structural Understanding Capabilities}

\begin{figure}[t]
  \centering
  \includegraphics[width=\linewidth]{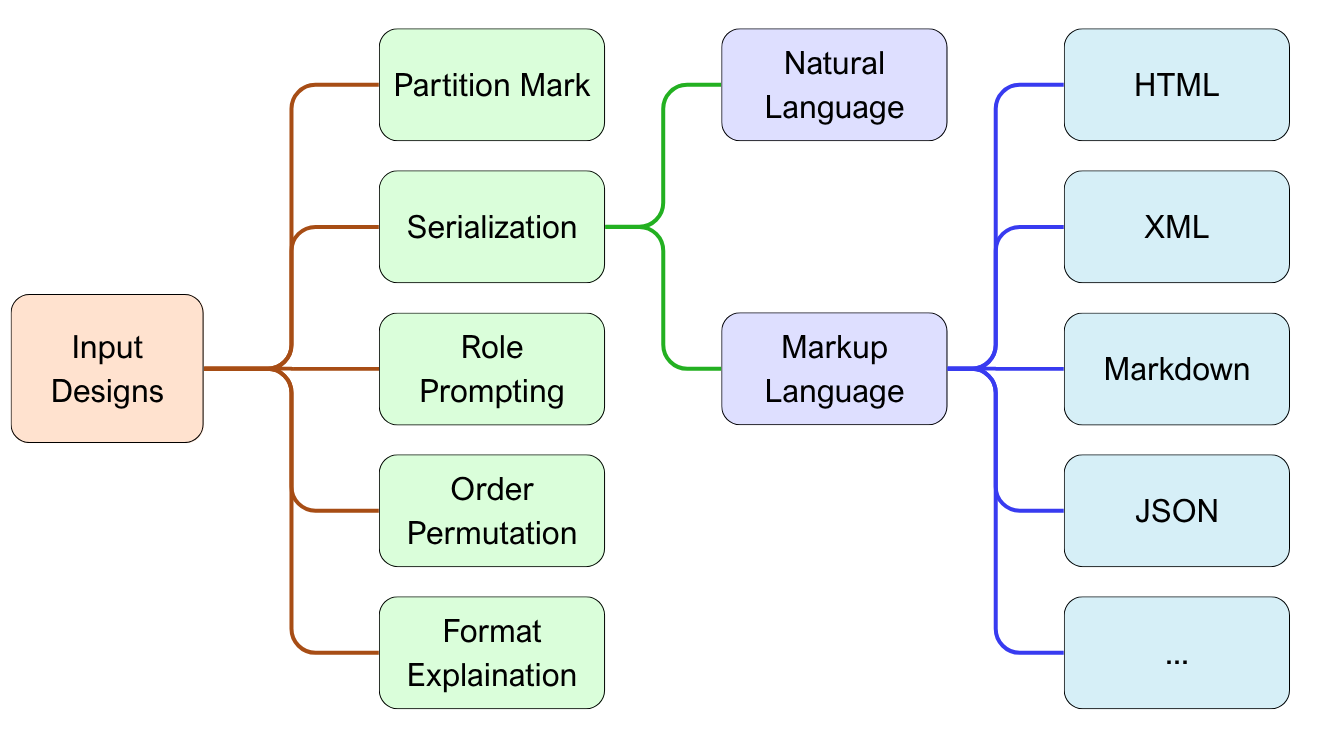}
  \caption{Input designs evaluated in the QASU benchmark. Each design is a combination of serialization format, layout choices, and prompt annotations.}
  \label{fig:input_designs}
\end{figure}

We categorize the fundamental abilities required to work effectively with questionnaire data into two complementary dimensions, illustrated in Figure~\ref{fig:input_designs}. These dimensions parallel the way human analysts approach data: first locating relevant information, then performing computations or logical reasoning over it.

\emph{1) Search \& Retrieval.}  
The first capability is finding information accurately in specific cells, rows, or columns of questionnaire data. This is similar to table QA, where even a simple value lookup means matching the question to the right parts of the table and handling complex structures \cite{herzig2020tapas}. In questionnaires, this means interpreting column headers, matching query terms to the appropriate variables, and identifying the correct answers without losing the context of the surrounding data. Diverse serialization formats, such as CSV, JSON, and XML, encode structure in very different ways, which can influence how easily an LLM can parse and navigate the data. A robust retrieval capability is therefore foundational: without it, downstream reasoning steps no matter how sophisticated cannot proceed reliably. QASU isolates this skill to assess how different input designs affect an LLM’s ability to recognize and extract the correct information across varying formats.

\emph{2) Aggregation \& Reasoning.}  
Besides locating specific entries, questionnaire analysis often needs to combine multiple pieces of information, applying filters, and performing computations to provide answers for more complex queries. This includes tasks where models must integrate several constraints, group related concepts, or apply logical and numerical operations across subsets of the data. Such operations are common in both human-led survey analysis and structured benchmarks like Spider \cite{yu-etal-2018-spider}, and they require the model to maintain a consistent mapping between the schema, the constraints, and the relevant rows. Crucially, these reasoning processes depend on accurate retrieval as an initial step; an error in locating the relevant data propagates directly into incorrect aggregate results. 


\subsection{Task Design}
\label{taskdesign}

\begin{figure}[t]
  \centering
  \includegraphics[width=\linewidth]{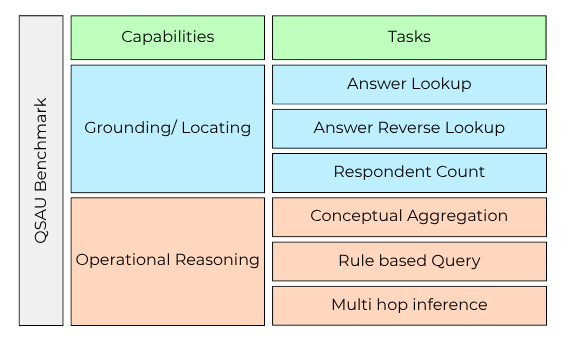}
  \caption{Task types in the QASU benchmark. Each task is designed to evaluate a specific structural skill in processing questionnaire data.}
  \label{fig:benchmark_tasks}
\end{figure}

At their core, the six tasks in QASU represent the fundamental operations behind any questionnaire analysis: retrieving specific answers, filtering by criteria, counting records, and combining conditions. These are the building blocks upon which more complex reasoning tasks, such as generating analytical reports, identifying statistical patterns, or conducting multi-step causal inference, must depend. We deliberately focus on operations that form the prerequisite for advanced analysis workflows. More specialized capabilities, such as handling missing-data imputation strategies, modeling temporal dynamics across longitudinal surveys, or navigating deeply hierarchical question structures with branching logic, are important but orthogonal challenges that introduce additional layers of complexity beyond structural understanding. By isolating these six foundational tasks, QASU provides a controlled environment to measure whether LLMs can reliably perform basic questionnaire operations before tackling higher-order analytical goals. Extensions to missing-data reasoning, temporal questionnaire analysis, and hierarchical skip-logic navigation are natural directions for future benchmarks once baseline structural competence has been established.

From this foundation, QASU evaluates structural understanding over serialized questionnaires rather than database tables. Let a serialized document be \(D = (Q, R)\), where \(Q\) maps question keys to their descriptors (for example, the text and option set for “Medical Condition”), and \(R = \{r_i\}\) is a set of respondent records. Each record \(r_i\) contains an identifier and an \texttt{answers} map from question keys to atomic values or codes. The concrete representation of \(D\) varies by format. In JSON it is a nested object with keys and values, in HTML it is a structured markup fragment, in Markdown it appears as a text table, and in XML it is a tag hierarchy. Across formats, the model must align a natural language query to a question key and then resolve the correct value or record by following the appropriate path in \(D\). An overview of benchmark tasks is illustrated in Figre~\ref{fig:benchmark_tasks}

\textit{Answer Lookup.}  
Given a respondent identifier \(i\) and a target question key \(k\), return the value \(v\) such that \(r_i.\texttt{answers}[k] = v\). This tests schema grounding and path resolution across formats. For JSON, this is a direct key access; for HTML or Markdown, it requires locating the field by header text and reading the associated entry in the respondent’s record.

\textit{Reverse Answer Lookup.}  
Given a question key \(k\) and a target value \(v\) that may be encoded as a letter for multiple choice, return the set of respondent identifiers \( \{ i : r_i.\texttt{answers}[k] = v \} \). This checks whether the model can search records by field value and, when applicable, map codes to their option labels using the information in \(Q\).

\textit{Respondent Count.}  
Return the number of respondent records \(|R|\), or the number that satisfy a stated global criterion when present. This requires detecting respondent boundaries in the serialized document and avoids double counting when formatting introduces repeated headers or decorative text.

\textit{Conceptual Aggregation.}  
Given a high level concept defined by a set of labels under a single question key, count the respondents whose answer falls in that set. For example, with \(k=\) “Medical Condition” and a concept defined as \{\textit{Diabetes}, \textit{Obesity}\}, return the total number of records whose \(\texttt{answers}[k]\) matches any member of the set, with code to label decoding when needed.

\textit{Multi hop Relational Inference.}  
Find the respondent IDs that meet all given conditions for different questions. This means matching values across fields and taking only the records that satisfy every condition. The task gets harder when there are more conditions and when answer codes need to be converted to their full text.

\textit{Rule based Querying.}  
Given a numerical predicate over a field, return the identifiers of respondents that satisfy the predicate. For example, with \(k=\) “Billing Amount” and a threshold, the model must parse numbers from text, apply the comparator, and report the matching records.

The tasks progress from direct retrieval to constrained reasoning. This ordering allows us to study how format and prompt choices affect alignment to question keys, robustness of path resolution across different serializations, and the reliability of composition over multiple constraints. Table~\ref{tab:qasu-input} lists the prompt templates used for each task.

\begin{table}[t]
\small
\centering
 \caption{Simple Query templates for the six structural tasks evaluated in the QASU benchmark}
  \label{tab:qasu-input}
\begin{adjustbox}{width=\columnwidth}
\begin{tabular}{ll}
\hline
\textbf{Task} & \textbf{Input} \\ \hline
Answer Lookup & \begin{tabular}[c]{@{}l@{}}What is the {[}attribute{]} of {[}respondent{]}?\\ Provide concisely only the attribute of\\ the respondent\end{tabular} \\ \hline
\begin{tabular}[c]{@{}l@{}}Answer Reverse\\ Lookup\end{tabular} & \begin{tabular}[c]{@{}l@{}}Which respondent has {[}attribute{]}? \\ Provide concisely only the respondent\\ number. Use "," to separate if there are\\ more than 1 respondent\end{tabular} \\ \hline
Respondent Count & \begin{tabular}[c]{@{}l@{}}How many respondents are there?\\  Provide concisely only the number\\  of respondents\end{tabular} \\ \hline
Conceptual Aggregation & \begin{tabular}[c]{@{}l@{}}How many respondents have {[}attribute{]}?\\ Provide concisely only the number of\\ respondents have that attribute\end{tabular} \\ \hline
\begin{tabular}[c]{@{}l@{}}Rule based\\ Querying\end{tabular} & \begin{tabular}[c]{@{}l@{}}Find all respondents matching {[}criteria{]}? \\ Provide concisely only the respondent\\  number. Use "," to separate if there are\\  more than 1 respondent\end{tabular} \\ \hline
\begin{tabular}[c]{@{}l@{}}Multi-hop relational\\ inference\end{tabular} & \begin{tabular}[c]{@{}l@{}}Which respondent has {[}attribute\_1{]}\\ and {[}attribute\_2{]}? Provide concisely \\ only the respondent number. Use "," to \\ separate if there are more than 1 respondent\end{tabular} \\ \hline
\end{tabular}
\end{adjustbox}
\end{table}

\subsection{Data collection and reformatting}
\label{datacollect}

\emph{Datasets.}
We collect questionnaire tables from five publicly available datasets spanning health, psychology, human–computer interaction, and software engineering. These include an anonymized hospital discharge file from Kaggle’s Healthcare Dataset \cite{prasad2020healthcare}, self-reported mental wellbeing responses from Mendeley Data \cite{topp2015mental}, the SUS-UTA7 corpus of usability scores for wearable systems \cite{mimbcd2022uta7}, the 2022 Stack Overflow Developer Survey \cite{stackoverflow2022survey}, and an ISBAR study assessing medical students’ hand-off performance. All sources are provided as CSV or XLSX files. We retain only the structural elements of the questionnaires, including question headers and respondent records. Entries with excessive missing values are discarded. While incomplete responses are common in real-world surveys, keeping them in the benchmark would make it harder to attribute errors to the model rather than to the absence of information. This filtering ensures that evaluation focuses on reasoning and comprehension rather than imputation.

Each dataset is reformatted into a unified representation that can be serialized into multiple formats, including JSON, HTML, XML, TTL, TXT, and Markdown. The questions component is where the question list and its type are stored, while the responses component records respondent id and their answers mapped to each question in question list. This harmonisation removes schema-specific variability and allows consistent downstream processing regardless of the chosen serialization.

For each benchmark instance, we randomly sample a number of respondents so that the serialized input fits within the token budget, while keeping all questions intact. A natural-language query is then appended, drawn from the templates in \S~\ref{taskdesign}. Each original dataset already contains diverse question types mixed together from multiple choice questions to open ended questions. The QASU benchmark is evaluated primarily in a one-shot setting, where the model sees one worked example from the same task type before answering. This design leverages evidence that few-shot prompting can elicit reasoning behaviors not observed in zero-shot settings \cite{brown2020gpt3}.

\emph{Leakage-resistant.}
To prevent potential data leakage from large language models (LLMs) that may have been trained on or have retrieval-augmented access to publicly available datasets, we apply a systematic obfuscation process before serialization. Without this step, models could recall memorized facts or entity associations from the original sources, artificially inflating benchmark performance and undermining the evaluation of reasoning ability rather than recall \cite{magar2022data, thakkar2021understanding}.

The obfuscation process operates as follows. First, respondent identifiers are shuffled and personally identifiable information such as human names and place names is encoded to anonymized placeholders. For numeric variables, we adopt a \emph{rank swapping} baseline \cite{domingo2001rank}: for each numeric variable $X$, records are first sorted by value; for each record at rank $r$, we randomly select a swap partner uniformly within the rank interval $[r-w, r+w]$, where the window $w$ is chosen as $5\%$ of $n$, the number of records. The values between these two records are then swapped. Rank swapping was selected because it preserves distributional properties (mean, variance, percentiles) critical for aggregation tasks while introducing sufficient perturbation to prevent exact record linkage. The $5\%$ window size follows established practice in statistical disclosure control for balancing privacy and utility. 

For multiple-choice (MCQ) variables, we first compute the empirical option distribution for each question across the dataset. Then, for each respondent's answer, we introduce controlled perturbation with probability $\alpha = 10\%$, replacing it with an alternative option sampled according to the global option distribution (e.g., if the distribution is 50\% A, 20\% B, 30\% C, replacements follow these proportions). This ensures that option frequencies remain realistic while preventing the model from exploiting memorized respondent–answer associations. This obfuscation strategy balances the need for realistic, structurally faithful questionnaire data with the requirement for a fair evaluation, isolating reasoning skills from memorization or retrieval of seen content.

\subsection{Evaluation}

We evaluate QASU using exact-match accuracy, defined as the proportion of questions where the model’s predicted answer matches the ground truth exactly after normalizing whitespace and letter case. This scoring choice follows established practice in benchmarks such as WikiSQL \cite{zhong2017wikisql} and TabFact \cite{chen2020tabfact}, and is well-suited here since all six QASU tasks have deterministic ground truths that require no paraphrase tolerance.

The benchmark is tested on six formats: JSON, XML, Markdown, HTML, Turtle, and plain text. Each format keeps a different level of structural information, helping us see whether mistakes come from poor reasoning or from missing layout information. This range also mirrors how questionnaires are stored in different fields.

In addition to format choice, we evaluate four prompt augmentations: (1) a brief syntax description of the serialized data, (2) a partition marker to visually separate sections, (3) a role label inspired by TaPEx \cite{liu2022tapex} that explicitly tags question and answer fields, and (4) a brief explanation of the serialization format. Combining these elements yields a rich input design space, shown in Figure~\ref{fig:input_designs}, with results in Table~\ref{tab:qasu-results}.

All prompts include explicit output constraints to standardize model behavior. 
Empirically, over 90\% of model outputs adhere to these instructions. During automatic scoring, we extract the predicted answer from each model output according to the predefined output format and compare it directly with the ground truth. This combination of strict output constraints, deterministic scoring, and automated answer extraction ensures that QASU provides a reliable and reproducible measure of LLM performance on questionnaire reasoning tasks.

\section{STRUCTURAL PROMPTING}

\begin{figure*}[t]
  \centering
  \includegraphics[width=0.85\linewidth]{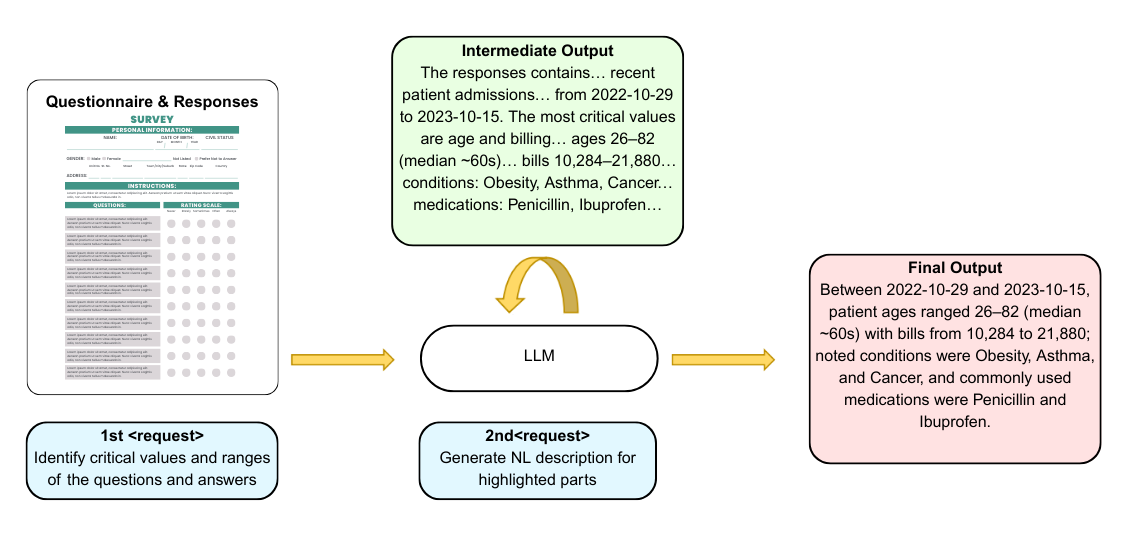}
  \caption{Illustration of self-augmented prompting. This process consists of two phases: 1) using self-augmented prompts to ask
the LLM to generate additional knowledge (intermediate output) about the table; 2) incorporating the self-augmented response
into the second prompt to request the final answer for a downstream task. As depicted in the figure, the LLM is able to identify
important values in the table, which assists in generating a more accurate answer for the downstream task.}
  \label{fig:self_aug}
\end{figure*}

Two consistent patterns emerge from our analyses of the QASU experiments (see \S~\ref{sec:experiments}): (1) models show a basic understanding of questionnaire structure but still make preventable mistakes on tasks like answer reverse lookups and respondent counting, and (2) the prompt phrasing and input format selection can change accuracy by a number of points without altering the underlying model.  These results are consistent with those of Sui et al. in their SUC benchmark for tables \cite{sui2024tablemeetsllm}, where performance was more affected by moving from JSON to HTML or adding a one-line format note than by improving the LLM itself. As in their work, we adopt \emph{self-augmented prompting} as an external method, and adapt it to the distinct characteristics of questionnaire data.

Self-augmented prompting operates in two sequential stages, as shown in Figure~\ref{fig:self_aug}. Rather than posing the final question directly, the first stage elicits \emph{structural probes} that explicitly surface critical schema facts before reasoning begins. In our adaptation, these probes may include identifying respondent ID patterns, enumerating question headers, or mapping value aliases for composite concepts (e.g., the full list of conditions under “metabolic condition”). The second stage integrates the output of the structural probe into a prompt containing the original task question, thereby anchoring the model’s reasoning in explicit structural knowledge rather than leaving it implicit. This approach is model-agnostic and avoids injecting handcrafted logic, relying entirely on the LLM’s own parsing of the serialised input.



Finally, we note that structural prompting is complementary to advances in context length or retrieval-augmented decoding. Even if future models eliminate truncation constraints, the two-stage process can still provide a clear scaffold that prevents the model from hallucinating nonexistent respondents, questions, or answer categories.

\section{EXPERIMENTS}
\label{sec:experiments}
\subsection{Experiment settings}

\textbf{Models.} We evaluate a range of contemporary large language models to assess the generalizability of our findings across different model families. In the main paper, we focus our analysis on two widely used models: \textbf{GPT-5-mini} and \textbf{Gemini-2.5-Flash}, which represent state-of-the-art commercial LLMs with strong performance on structured reasoning tasks. Unless otherwise stated, all completions are generated with temperature set to 0 and one output per prompt. This setup follows the deterministic setting used in earlier research on table reasoning and text-to-SQL \cite{brown2020gpt3}, which means that any changes in performance are due to how the input was designed, not random sampling. By fixing the decoding parameters, we can directly link changes in accuracy to structural prompts or serialization format, not to randomness in the generation process. To demonstrate broader applicability, we also evaluated three additional models spanning both open-source and commercial offerings (Qwen3-32B, Llama3-70B, and Amazon Nova Lite); these comprehensive results across all models are presented in Appendix~\ref{appendix:extended-results}.

\textbf{Datasets and query pool.} To examine whether the structural capabilities measured by QASU generalize across domains, we compile a diverse evaluation set from five publicly available questionnaire datasets covering health, psychology, software engineering, and clinical training. These include: (1) an anonymized hospital discharge dataset from \textbf{Kaggle’s Healthcare Dataset} \cite{prasad2020healthcare}; (2) \textbf{Self-reported mental well-being responses} hosted on Mendeley Data \cite{topp2015mental}; (3) the \textbf{SUS-UTA7} usability study for wearable systems \cite{mimbcd2022uta7}; (4) the \textbf{2022 Stack Overflow Developer Survey} \cite{stackoverflow2022survey}; and (5) an \textbf{ISBAR} dataset rating medical students’ patient hand-off performance. 

For each dataset, we generate independent benchmark cases per QASU task, where each case is a random sample of respondents fitted to the token budget, paired with one of the six task templates from Table~\ref{tab:qasu-input}. Each case is serialized into all six data formats, producing a balanced evaluation across tasks, datasets, and serialization approaches. In question list of each dataset, there are mixed types of questionnaire data such as Likert scales, multiple-choice grids, and open-ended text responses, preventing a model from memorizing or exploiting a single questionnaire schema.

\textbf{Evaluation metric.}We check the correctness of all jobs by comparing them to the exact match, which is calculated after normalizing the white space and letter case. This decision is consistent with known protocols in structured data reasoning benchmarks like WikiSQL and TabFact \cite{zhong2017wikisql,chen2020tabfact}, where deterministic ground truths eliminate the necessity for semantic similarity scoring. To reduce confusion caused by the format, each prompt includes an explicit output-format instruction emphasizing that answers must be human-readable values. It also instructs that letter codes from multiple-choice questions must be matched to their corresponding option text. This limitation, along with deterministic decoding, makes it possible to compare serialization formats and prompting variants in a clear and repeatable way.

\subsection{Results}

\begin{table*}[t]
\centering
\caption{Results of the benchmark on GPT-5-mini and Gemini-2.5-Flash. 'GPT' refers to GPT-5-mini and 'Gemini' refers to Gemini-2.5-Flash. Each column uses a graded color scale where deeper colors signify superior performance. For comprehensive results including three additional models (Qwen3-32B, Llama3-70B, Amazon Nova Lite), see Table~\ref{tab:all-tasks-paired} in Appendix~\ref{appendix:extended-results}.}
\label{tab:qasu-results}
\resizebox{0.95\textwidth}{!}{
\begin{tabular}{l rr rr rr rr rr rr}
\toprule
& \multicolumn{2}{c}{Answer Lookup} & \multicolumn{2}{c}{Reverse Lookup} & \multicolumn{2}{c}{Resp. Count} & \multicolumn{2}{c}{Concept. Agg.} & \multicolumn{2}{c}{Rule Query} & \multicolumn{2}{c}{Multi hop} \\
\cmidrule(lr){2-3} \cmidrule(lr){4-5} \cmidrule(lr){6-7} \cmidrule(lr){8-9} \cmidrule(lr){10-11} \cmidrule(lr){12-13}
Format & GPT & Gemini & GPT & Gemini & GPT & Gemini & GPT & Gemini & GPT & Gemini & GPT & Gemini \\
\midrule
HTML& \cellcolor{mygreen4}{\textbf{75.4\%}} & \cellcolor{myblue3}{86.9\%} & \cellcolor{mygreen4}{\textbf{88.0\%}} & \cellcolor{myblue4}{\textbf{80.6\%}} & \cellcolor{mygreen2}{94.9\%} & \cellcolor{myblue1}{93.7\%} & \cellcolor{mygreen4}{\textbf{96.6\%}} & \cellcolor{myblue2}{94.9\%} & \cellcolor{mygreen3}{94.3\%} & \cellcolor{myblue1}{91.4\%} & \cellcolor{mygreen4}{\textbf{72.6\%}} & \cellcolor{myblue1}{71.4\%} \\
JSON& \cellcolor{mygreen3}{72.0\%} & \cellcolor{myblue2}{84.6\%} & \cellcolor{mygreen3}{86.9\%} & \cellcolor{myblue3}{80.0\%} & \cellcolor{mygreen4}{\textbf{96.0\%}} & \cellcolor{myblue2}{94.9\%} & \cellcolor{mygreen3}{96.0\%} & \cellcolor{myblue4}{\textbf{96.0\%}} & \cellcolor{mygreen1}{92.6\%} & \cellcolor{myblue2}{92.0\%} & \cellcolor{mygreen1}{66.3\%} & \cellcolor{myblue2}{73.7\%} \\
MD& \cellcolor{mygreen3}{72.0\%} & \cellcolor{myblue2}{84.6\%} & \cellcolor{mygreen3}{86.3\%} & \cellcolor{myblue2}{78.3\%} & \cellcolor{mygreen2}{94.9\%} & \cellcolor{myblue4}{\textbf{96.0\%}} & \cellcolor{mygreen2}{95.4\%} & \cellcolor{myblue1}{93.7\%} & \cellcolor{mygreen2}{93.7\%} & \cellcolor{myblue3}{92.6\%} & \cellcolor{mygreen3}{72.0\%} & \cellcolor{myblue3}{75.4\%} \\
TTL & \cellcolor{mygreen1}{66.6\%} & \cellcolor{myblue3}{86.3\%} & \cellcolor{mygreen1}{85.1\%} & \cellcolor{myblue1}{76.6\%} & \cellcolor{mygreen1}{94.3\%} & \cellcolor{myblue3}{95.4\%} & \cellcolor{mygreen4}{\textbf{96.6\%}} & \cellcolor{myblue3}{95.4\%} & \cellcolor{mygreen4}{\textbf{95.4\%}} & \cellcolor{myblue4}{\textbf{93.7\%}} & \cellcolor{mygreen1}{67.4\%} & \cellcolor{myblue2}{74.3\%} \\
TXT & \cellcolor{mygreen2}{70.9\%} & \cellcolor{myblue4}{\textbf{88.0\%}} & \cellcolor{mygreen2}{85.7\%} & \cellcolor{myblue3}{80.0\%} & \cellcolor{mygreen1}{94.3\%} & \cellcolor{myblue4}{\textbf{96.0\%}} & \cellcolor{mygreen2}{95.4\%} & \cellcolor{myblue1}{93.1\%} & \cellcolor{mygreen3}{94.3\%} & \cellcolor{myblue1}{90.9\%} & \cellcolor{mygreen2}{69.1\%} & \cellcolor{myblue4}{\textbf{78.3\%}} \\
XML & \cellcolor{mygreen3}{72.0\%} & \cellcolor{myblue1}{77.7\%} & \cellcolor{mygreen3}{86.3\%} & \cellcolor{myblue2}{78.3\%} & \cellcolor{mygreen3}{95.4\%} & \cellcolor{myblue1}{93.1\%} & \cellcolor{mygreen2}{95.4\%} & \cellcolor{myblue2}{94.3\%} & \cellcolor{mygreen2}{93.7\%} & \cellcolor{myblue1}{88.6\%} & \cellcolor{mygreen2}{68.6\%} & \cellcolor{myblue1}{68.0\%} \\
\bottomrule
\end{tabular}%
}
\end{table*}

\begin{table*}[ht]
\centering
\caption{Study of different input design choices against the HTML baseline for GPT-5-mini and Gemini-2.5-Flash. For ablation results on additional models, see Table~\ref{tab:qasu-results-input-des-full} in Appendix~\ref{appendix:extended-results}.}
\label{tab:qasu-results-input-des}
\resizebox{0.95\textwidth}{!}{%
\begin{tabular}{l rr rr rr rr rr rr}
\toprule
& \multicolumn{2}{c}{Answer Lookup} & \multicolumn{2}{c}{Reverse Lookup} & \multicolumn{2}{c}{Resp. Count} & \multicolumn{2}{c}{Concept. Agg.} & \multicolumn{2}{c}{Rule Query} & \multicolumn{2}{c}{Multi hop} \\
\cmidrule(lr){2-3} \cmidrule(lr){4-5} \cmidrule(lr){6-7} \cmidrule(lr){8-9} \cmidrule(lr){10-11} \cmidrule(lr){12-13}
Input Design & Acc & $\Delta$ & Acc & $\Delta$ & Acc & $\Delta$ & Acc & $\Delta$ & Acc & $\Delta$ & Acc & $\Delta$ \\
\midrule
\multicolumn{13}{l}{\textbf{GPT-5-mini}} \\
\midrule
Markup Lang. HTML & 75.4\% & 0.0\% & 88.0\% & 0.0\% & 94.9\% & 0.0\% & 96.6\% & 0.0\% & 94.3\% & 0.0\% & 72.6\% & 0.0\% \\
w/o format explanation & 74.3\% & \cellcolor{myred1}{-1.1\%} & 84.0\% & \cellcolor{myred1}{-4.0\%} & 92.6\% & \cellcolor{myred1}{-2.3\%} & 93.1\% & \cellcolor{myred1}{-3.5\%} & 87.4\% & \cellcolor{myred1}{-6.9\%} & 72.0\% & \cellcolor{myred1}{-0.6\%} \\
w/o partition mark & 66.9\% & \cellcolor{myred2}{-8.5\%} & 72.0\% & \cellcolor{myred3}{-16.0\%} & 86.3\% & \cellcolor{myred2}{-8.6\%} & 89.1\% & \cellcolor{myred1}{-7.5\%} & 80.0\% & \cellcolor{myred2}{-14.3\%} & 56.0\% & \cellcolor{myred3}{-16.6\%} \\
w/o role prompting & 75.4\% & 0.0\% & 79.4\% & \cellcolor{myred2}{-8.6\%} & 90.3\% & \cellcolor{myred1}{-4.6\%} & 92.6\% & \cellcolor{myred1}{-4.0\%} & 81.7\% & \cellcolor{myred2}{-12.6\%} & 69.9\% & \cellcolor{myred1}{-5.7\%} \\
w/o change order & 64.0\% & \cellcolor{myred2}{-11.4\%} & 60.0\% & \cellcolor{myred4}{-28.0\%} & 81.1\% & \cellcolor{myred2}{-13.8\%} & 77.7\% & \cellcolor{myred3}{-18.9\%} & 64.0\% & \cellcolor{myred4}{-30.3\%} & 49.1\% & \cellcolor{myred4}{-23.5\%} \\
w/o 1-shot & 65.1\% & \cellcolor{myred2}{-10.3\%} & 77.1\% & \cellcolor{myred2}{-10.9\%} & 88.6\% & \cellcolor{myred1}{-6.3\%} & 88.6\% & \cellcolor{myred2}{-8.0\%} & 79.4\% & \cellcolor{myred2}{-14.9\%} & 47.4\% & \cellcolor{myred4}{-25.2\%} \\
\midrule
\multicolumn{13}{l}{\textbf{Gemini-2.5-Flash}} \\
\midrule
Markup Lang. HTML & 86.9\% & 0.0\% & 80.6\% & 0.0\% & 93.7\% & 0.0\% & 94.9\% & 0.0\% & 91.4\% & 0.0\% & 71.4\% & 0.0\% \\
w/o format explanation & 81.7\% & \cellcolor{myred1}{-5.2\%} & 80.0\% & \cellcolor{myred1}{-0.6\%} & 93.2\% & \cellcolor{myred1}{-0.5\%} & 94.9\% & 0.0\% & 91.1\% & \cellcolor{myred1}{-0.3\%} & 74.3\% & \cellcolor{myblue1}{+2.9\%} \\
w/o partition mark & 62.3\% & \cellcolor{myred4}{-24.6\%} & 60.6\% & \cellcolor{myred4}{-20.0\%} & 75.4\% & \cellcolor{myred3}{-18.3\%} & 71.1\% & \cellcolor{myred4}{-23.8\%} & 75.7\% & \cellcolor{myred3}{-15.7\%} & 57.1\% & \cellcolor{myred3}{-14.3\%} \\
w/o role prompting & 82.9\% & \cellcolor{myred1}{-4.0\%} & 81.1\% & \cellcolor{myblue1}{+0.5\%} & 92.0\% & \cellcolor{myred1}{-1.7\%} & 96.0\% & \cellcolor{myblue1}{+1.1\%} & 86.9\% & \cellcolor{myred1}{-4.5\%} & 71.2\% & \cellcolor{myred1}{-0.2\%} \\
w/o change order & 82.9\% & \cellcolor{myred1}{-4.0\%} & 74.9\% & \cellcolor{myred1}{-5.7\%} & 96.6\% & \cellcolor{myblue1}{+2.9\%} & 93.7\% & \cellcolor{myred1}{-1.2\%} & 82.3\% & \cellcolor{myred2}{-9.1\%} & 68.6\% & \cellcolor{myred1}{-2.8\%} \\
w/o 1-shot & 74.9\% & \cellcolor{myred2}{-12.0\%} & 83.4\% & \cellcolor{myblue1}{+2.8\%} & 96.0\% & \cellcolor{myblue1}{+2.3\%} & 93.1\% & \cellcolor{myred1}{-1.8\%} & 90.9\% & \cellcolor{myred1}{-0.5\%} & 65.1\% & \cellcolor{myred1}{-6.3\%} \\
\bottomrule
\end{tabular}%
}
\end{table*}

\emph{Benchmark Highlights}
When comparing six different input formats across our benchmark on GPT-5-mini and Gemini-2.5-Flash, it becomes clear that there is no single ``best'' format for every type of reasoning, as shown in Table~\ref{tab:qasu-results}. Different formats excel at different tasks, indicating that format choice can meaningfully influence model performance. These trends hold consistently across a broader range of LLMs, including open-source models (see Appendix~\ref{appendix:extended-results} for comprehensive results with Qwen3-32B, Llama3-70B, and Amazon Nova Lite).

Lastly, we compare standard 1-shot prompts with self-augmented prompts, where the model first generates its own auxiliary notes or interpretations before producing the final answer. Across all datasets, self-augmentation generally matched or exceeded 1-shot performance, in some cases by a considerable margin (Table~\ref{tab:self-aug-performance}). The largest gains occurred when the model explicitly identified critical values and ranges or provided a self-structural information description, effectively producing a tailored ``cheat sheet'' from the input. 

For GPT-5-mini, these strategies yielded notable improvements on datasets with diverse question types, such as Mental-health (+3.1\% over 1-shot) and SO-2022 (+1.7\%). Gemini-2.5-Flash showed more modest but still consistent benefits from self-augmentation, particularly in the \textsc{ISBAR} and \textsc{SUS-UTA7} datasets. 

Interestingly, the self format explanation option delivered strong results in GPT-5-mini, but was less effective for Gemini, hinting that some models may gain more from structural restatement while others already leverage structural cues without needing explicit articulation. Overall, these results suggest that letting the model pre-process the input in its own words can help it adapt better to dataset-specific nuances than rigid, human-crafted instructions.

\newpage

\paragraph{Natural language vs. markup language.} For both GPT-5-mini and Gemini-2.5-Flash, plain-text (TXT) representation performed competitively with structured markup formats such as HTML, JSON, and XML. For certain reasoning types, especially Multi-hop and Concept Aggregation, both models benefited from the familiarity and flexibility of a narrative style. In contrast, markup formats generally excelled in more rigid tasks such as Response Counting and Answer Reverse Lookup, where explicit structure is advantageous.

\paragraph{One-shot vs. zero-shot prompting.} Across the benchmark, one-shot prompting consistently outperformed zero-shot prompting for both evaluated models. On average, one-shot scores were higher by a noticeable margin across all reasoning types, confirming prior findings that even a single worked example can prime LLMs to adopt more effective reasoning strategies. This effect was especially pronounced in tasks requiring multi-step reasoning, where seeing an example helped both GPT-5-mini and Gemini-2.5-Flash better chain intermediate steps to reach the correct answer.

\emph{Self-Augmented vs.\ 1-shot Prompting}
\label{sec:self-aug-vs-1shot} Lastly, we compare standard 1-shot prompts with self-augmented prompts, as illustrated in Table~\ref{tab:qasu-results-input-des} where the model first generates its own auxiliary notes or interpretations before producing the final answer. Across all datasets, self-augmentation generally matched or exceeded 1-shot performance, in some cases by a considerable margin (Table~\ref{tab:self-aug-performance}). The largest gains occurred when the model explicitly identified critical values and ranges or provided a structural information description, effectively producing a tailored ``cheat sheet'' from the input. 

For GPT-5-mini, these strategies yielded notable improvements on datasets with diverse question types, such as \textsc{Mental-health} (+3.1\% over 1-shot) and \textsc{SO-2022} (+1.7\%). Gemini-2.5-Flash showed more modest but still consistent benefits from self-augmentation, particularly in the \textsc{ISBAR} and \textsc{SUS-UTA7} datasets. 

Interestingly, the self format explanation option delivered strong results in GPT-5-mini, but was less effective for Gemini, hinting that some models may gain more from structural restatement while others already leverage structural cues without needing explicit articulation. Overall, these results suggest that letting the model pre-process the input in its own words can help it adapt better to dataset-specific nuances than rigid, human-crafted instructions.

\section{RELATED WORK}

\begin{table*}[ht]
\centering
\caption{Comparison of self-augmented and 1-shot prompting across various design variants on downstream tasks for GPT-5-mini and Gemini-2.5-Flash. Refer to Table~\ref{tab:self_aug_prompts} for details on different self-augmented prompting options. "SA" stands for "Self-augmented". For results on additional models, see Table~\ref{tab:self-aug-performance-full} in Appendix~\ref{appendix:extended-results}.}
\label{tab:self-aug-performance}
\resizebox{0.93\textwidth}{!}{%
\begin{tabular}{llccccc}
\toprule
\multirow{2}{*}{\textbf{Type}} & \multirow{2}{*}{\textbf{Choice}}
& \textbf{Healthcare} & \textbf{ISBAR} & \textbf{Mental-health} & \textbf{SO-2022} & \textbf{SUS-UTA7} \\
\cmidrule(lr){3-7}
& & \textbf{Acc} & \textbf{Acc} & \textbf{Acc} & \textbf{Acc} & \textbf{Acc} \\
\midrule
\multicolumn{7}{l}{\textbf{GPT 5 Mini}} \\
\midrule
1-shot & 1-shot & \cellcolor{myblue2}{45.0\%} & \cellcolor{myblue2}{68.0\%} & \cellcolor{myblue4}{\textbf{85.0\%}} & \cellcolor{myblue4}{\textbf{84.0\%}} & \cellcolor{myblue4}{\textbf{95.0\%}} \\
1-shot & w/o change order & \cellcolor{myblue3}{47.0\%} & \cellcolor{myblue1}{60.0\%} & \cellcolor{myblue1}{62.0\%} & \cellcolor{myblue2}{66.0\%} & \cellcolor{myblue3}{85.0\%} \\
1-shot & w/o format explanation & \cellcolor{myblue4}{\textbf{50.0\%}} & \cellcolor{myblue1}{67.0\%} & \cellcolor{myblue3}{78.0\%} & \cellcolor{myblue4}{81.0\%} & \cellcolor{myblue3}{84.0\%} \\
1-shot & w/o partition mark & \cellcolor{myblue2}{46.0\%} & \cellcolor{myblue3}{70.0\%} & \cellcolor{myblue3}{74.0\%} & \cellcolor{myblue3}{75.0\%} & \cellcolor{myblue2}{83.0\%} \\
1-shot & w/o role prompting & \cellcolor{myblue3}{48.0\%} & \cellcolor{myblue2}{68.0\%} & \cellcolor{myblue4}{80.0\%} & \cellcolor{myblue3}{79.0\%} & \cellcolor{myblue4}{88.0\%} \\
\midrule
SA & self format explanation & \cellcolor{myblue4}{49.0\%} & \cellcolor{myblue4}{\textbf{72.0\%}} & \cellcolor{myblue2}{72.0\%} & \cellcolor{myblue2}{70.0\%} & \cellcolor{myblue1}{73.0\%} \\
SA & self critical values and ranges identification & \cellcolor{myblue1}{30.0\%} & \cellcolor{myblue3}{71.0\%} & \cellcolor{myblue1}{57.0\%} & \cellcolor{myblue1}{58.0\%} & \cellcolor{myblue2}{77.0\%} \\
SA & self structural information description & \cellcolor{myblue1}{42.0\%} & \cellcolor{myblue4}{\textbf{72.0\%}} & \cellcolor{myblue2}{68.0\%} & \cellcolor{myblue1}{66.0\%} & \cellcolor{myblue1}{70.0\%} \\
\midrule
\multicolumn{7}{l}{\textbf{Gemini 2.5 Flash}} \\
\midrule
1-shot & 1-shot & \cellcolor{myblue4}{50.0\%} & \cellcolor{myblue3}{82.0\%} & \cellcolor{myblue4}{\textbf{95.0\%}} & \cellcolor{myblue4}{\textbf{83.0\%}} & \cellcolor{myblue3}{91.0\%} \\
1-shot & w/o change order & \cellcolor{myblue2}{42.0\%} & \cellcolor{myblue1}{70.0\%} & \cellcolor{myblue2}{89.0\%} & \cellcolor{myblue2}{79.0\%} & \cellcolor{myblue2}{90.0\%} \\
1-shot & w/o format explanation & \cellcolor{myblue3}{45.0\%} & \cellcolor{myblue2}{73.0\%} & \cellcolor{myblue3}{89.0\%} & \cellcolor{myblue4}{\textbf{83.0\%}} & \cellcolor{myblue4}{\textbf{94.0\%}} \\
1-shot & w/o partition mark & \cellcolor{myblue2}{44.0\%} & \cellcolor{myblue1}{63.0\%} & \cellcolor{myblue2}{80.0\%} & \cellcolor{myblue2}{74.0\%} & \cellcolor{myblue3}{90.0\%} \\
1-shot & w/o role prompting & \cellcolor{myblue1}{40.0\%} & \cellcolor{myblue1}{70.0\%} & \cellcolor{myblue4}{92.0\%} & \cellcolor{myblue3}{80.0\%} & \cellcolor{myblue4}{92.0\%} \\
\midrule
SA & self format explanation & \cellcolor{myblue1}{35.0\%} & \cellcolor{myblue4}{83.0\%} & \cellcolor{myblue3}{90.0\%} & \cellcolor{myblue3}{82.0\%} & \cellcolor{myblue2}{63.0\%} \\
SA & self critical values and ranges identification & \cellcolor{myblue4}{\textbf{70.0\%}} & \cellcolor{myblue4}{\textbf{91.0\%}} & \cellcolor{myblue1}{58.0\%} & \cellcolor{myblue1}{58.0\%} & \cellcolor{myblue1}{60.0\%} \\
SA & self structural information description & \cellcolor{myblue3}{45.0\%} & \cellcolor{myblue3}{78.0\%} & \cellcolor{myblue1}{77.0\%} & \cellcolor{myblue1}{72.0\%} & \cellcolor{myblue1}{58.0\%} \\
\bottomrule
\end{tabular}%
}
\end{table*}

\begin{table*}[ht]
\centering
\caption{Self-augmented prompting instructions.}
\label{tab:self_aug_prompts}
\resizebox{0.95\textwidth}{!}{%
\begin{tabular}{l p{0.7\textwidth}}
\toprule
\textbf{Method} & \textbf{Instruction} \\
\midrule
Format explanation & Generate short format specification and description of the survey within five sentences. \\
\midrule
Critical values and ranges identification & Identify critical values and ranges of the survey related within five sentences. \\
\midrule
Structural information description & Describe structural information, patterns and statistics of the survey within five sentences. \\
\bottomrule
\end{tabular}}
\end{table*}

\textbf{LLMs and structured data.}
Large language models have been increasingly used to deal with tasks that involve structured inputs like databases, forms, and semi-structured text. 
Before this, there are common approaches combined natural-language understanding with symbolic execution, helping to translate questions into SQL queries or logical forms \cite{zhong2017wikisql,yu-etal-2018-spider}.
Later approaches improved results by pretraining models on large collections of structured data \cite{herzig2020tapas,liu2022tapex}.

Other researchers have created artificial datasets to help models handle a wider variety of data formats \cite{yu2021grappa}, set up extra rules during the answer-generation process to ensure the outputs are always valid \cite{scholak2021picard}, and mixed structured data with regular text to improve how well models can reason \cite{jiang2023structgpt}.
Recent studies also show that careful prompt design can boost model performance, even in situations where the model has not seen similar tasks before. Research such as Zero-Shot Chain-of-Thought and Table Meets LLM demonstrates that careful wording, formatting, and use of delimiters can make LLMs much better at working with structured data \cite{kojima2022zeroshotcot,sui2024tablemeetsllm}.
Our benchmark builds on these findings but focuses on questionnaires, which often have more complex relationships within each record and a wider variety of question types than typical database tasks.

\textbf{Prompt engineering and self-augmentation.}
Recent prompting strategies, such as chain-of-thought reasoning \cite{wei2022cot} and self-consistency \cite{wang2023sc}, encourage large language models (LLMs) to articulate their reasoning step by step. However, these models can still face difficulties when tasks demand highly precise symbolic references. Some techniques, like Zero-Shot-CoT \cite{kojima2022zeroshotcot}, work by giving the model a short phrase that encourages it to explain its thinking step by step. Least-to-most prompting \cite{zhou2023leasttomost} helps by breaking down complex problems into a series of smaller questions. On top of that, instruction tuning \cite{wei2022flan} helps models follow instructions more accurately across many different types of tasks.
Building on these ideas, Sui et al.\ introduced \textit{self-augmented prompting}, where the model first generates explicit structural hints and then uses them to answer the question \cite{sui2024tablemeetsllm}.
We adapt this approach to questionnaire grids and observe a consistent improvement of about 3\% over direct prompting, without adding any handcrafted logic.
For evaluation considerations complementary to our setup, retrieval-aware generation metrics have also been explored in recent work \cite{li2025retrievaleval}.


\textbf{Questionnaire Standards and Interoperability.}
There is no single agreed-upon standard for representing questionnaires, which creates challenges for both sharing and automated analysis.
The DDI Codebook is common in social-science archives \cite{vardigan2008ddi}, FHIR JSON is required for many clinical forms, and the Survey Ontology encodes forms in linked data triples \cite{scandolari2021surveyontology}.
A recent study of German health research identified thirty-two semantic standards and seven syntactic standards in active use \cite{vorisek2024towards}.
Our results show that this lack of consistency has a measurable impact: for example, HTML with embedded format notes achieves 6\% higher accuracy than raw JSON, likely because it conveys richer structural information to the model; similar sensitivity to input design has also been documented in studies of structure-aware LLMs \cite{sui2024tablemeetsllm}, and structure-aware generation continues to be explored in applied domains \cite{chen2025medtranstab}.

\section{CONCLUSION}

This work introduces QASU, the first benchmark dedicated to probing large language models' structural understanding of questionnaire data. By systematically varying serialization formats, prompt augmentations, and evaluation settings across multiple contemporary LLMs, QASU isolates the role of input design in tasks that span answer lookup, answer reverse lookup, aggregation, and multi-hop reasoning. Our findings show that, much like in table-based reasoning, LLMs exhibit a basic grasp of questionnaire structure but remain sensitive to format cues and explicit schema hints. In particular, the gap between the best and worst-performing designs can exceed six accuracy points, underscoring the importance of careful input preparation.

Through evaluation on GPT-5-mini and Gemini-2.5-Flash as primary models, with additional validation on open-source alternatives (detailed in Appendix~\ref{appendix:extended-results}), we demonstrate that these patterns hold across different model families. We build on earlier work on structural prompting \cite{sui2024tablemeetsllm} and adapt self-augmented prompting to questionnaire grids, observing consistent gains without having to write any logic by hand. This shows that prompting tactics that were first devised for tabular data can also work well in surveys. In addition to benchmarking, the insights from QASU provide practical advice for applied survey analysis: selecting a more detailed serialization and including light structural annotations can greatly enhance model accuracy at little to no expense.

We believe that QASU will serve as a useful benchmark for evaluating future LLMs and provide actionable guidance for researchers who engage with survey data in health, social science, and beyond. By bridging the gap between academic standards and the realities of different types of questionnaires, QASU helps us better understand how large models deal with one of the most ubiquitous but least studied types of structured data.

\section*{ETHICAL CONSIDERATIONS}
Questionnaire data frequently contains sensitive or personally identifiable information, such as health records, demographic profiles, or workplace feedback. Even when identifiers are removed, indirect re-identification remains possible through unique combinations of answers. To mitigate these risks, all source datasets in QASU are drawn from public releases that have been anonymized by their original curators, and we discard entries with excessive missing values to avoid leaking potentially identifying patterns in response sparsity. No manual inspection of raw identifiable content is performed beyond automated schema validation.

A further consideration is the cultural and linguistic scope of our benchmark. The current QASU release focuses exclusively on English-language questionnaires, most of which originate from Western healthcare, usability, and software-engineering contexts. Extending QASU to cover multilingual or cross-cultural settings would require additional safeguards, including culturally sensitive schema interpretation and bias detection. Moreover, questionnaires designed for non-Western populations often employ different response scales, question structures, and cultural assumptions that may not transfer directly to our evaluation framework.

The evaluation protocol assumes that all null or missing values in the data are excluded from the reasoning target. This guarantees fairness in assessing structural comprehension by ensuring that errors stem from model limitations rather than ambiguity in the ground truth. However, this assumption contrasts with certain real-world survey analysis situations where managing missing data is essential to the task. In practice, missing values may carry information about respondent behavior, survey design, or data quality issues. Our obfuscation procedures also introduce a controlled amount of noise that, while necessary for leakage prevention, may slightly reduce the fidelity of the benchmark relative to pristine real-world data. Researchers applying QASU insights to downstream systems should carefully reassess these assumptions in operational contexts where missing values may themselves be informative and where data have not undergone privacy-preserving transformations.

We believe that responsible benchmarking on questionnaire data requires not only technical rigor in measuring model capabilities, but also careful attention to privacy, cultural representativeness, and task realism. Addressing these aspects is essential for any future expansion of QASU toward a broader and more diverse set of structured-data reasoning challenges.


\bibliographystyle{ACM-Reference-Format}
\bibliography{sample-base}

\appendix
\section{Extended Results with Additional Models}
\label{appendix:extended-results}

This appendix presents comprehensive evaluation results for three additional large language models beyond GPT-5-mini and Gemini-2.5-Flash reported in the main paper: Qwen3-32B, Llama3-70B, and Amazon Nova Lite. These results demonstrate the generalizability of our findings across a broader range of contemporary LLMs, including both commercial and open-source models.

\subsection{Comprehensive Benchmark Results}

Table~\ref{tab:all-tasks-paired} presents the complete performance breakdown across all six QASU tasks and five models. The results reveal several notable patterns. First, GPT-5-mini maintains the strongest overall performance, particularly on complex reasoning tasks like Multi-hop inference. Second, open-source models (Qwen3-32B and Llama3-70B) show competitive performance on structured retrieval tasks but lag significantly on aggregation and reasoning tasks. Third, Amazon Nova Lite, despite being optimized for efficiency, achieves respectable performance on basic lookup tasks but struggles with multi-step reasoning.

\begin{table*}[t]
\centering
\caption{Comprehensive benchmark results for all tasks. 'GPT' refers to GPT-5-mini, 'Gemini' to Gemini-2.5-Flash, while 'Qwen3', 'Llama3', and 'Nova' correspond to Qwen3-32B, Llama3-70B, and Amazon Nova Lite, respectively. Each cell's color is scaled relative to other formats within the same task and model, where deeper colors signify superior performance.}
\label{tab:all-tasks-paired}

\resizebox{0.9\textwidth}{!}{
\begin{tabular}{l rrrrr rrrrr}
\toprule
& \multicolumn{5}{c}{Answer Lookup} & \multicolumn{5}{c}{Reverse Lookup} \\
\cmidrule(lr){2-6} \cmidrule(lr){7-11}
Format & GPT & Gemini & Qwen3 & Llama3 & Nova & GPT & Gemini & Qwen3 & Llama3 & Nova \\
\midrule
HTML     & \cellcolor{mygreen4}{\textbf{75.4\%}} & \cellcolor{myblue3}{86.9\%} & \cellcolor{mygreen1}{67.6\%} & \cellcolor{myblue1}{64.6\%} & \cellcolor{mygreen1}{57.1\%} & \cellcolor{mygreen4}{\textbf{88.0\%}} & \cellcolor{myblue4}{\textbf{80.6\%}} & \cellcolor{mygreen1}{50.1\%} & \cellcolor{myblue1}{52.7\%} & \cellcolor{mygreen1}{45.3\%} \\
JSON     & \cellcolor{mygreen3}{72.0\%} & \cellcolor{myblue2}{84.6\%} & \cellcolor{mygreen3}{75.4\%} & \cellcolor{myblue4}{80.6\%} & \cellcolor{mygreen2}{57.7\%} & \cellcolor{mygreen3}{86.9\%} & \cellcolor{myblue3}{80.0\%} & \cellcolor{mygreen2}{51.7\%} & \cellcolor{myblue3}{57.4\%} & \cellcolor{mygreen3}{48.4\%} \\
MD       & \cellcolor{mygreen3}{72.0\%} & \cellcolor{myblue2}{84.6\%} & \cellcolor{mygreen2}{73.1\%} & \cellcolor{myblue3}{78.3\%} & \cellcolor{mygreen4}{62.3\%} & \cellcolor{mygreen3}{86.3\%} & \cellcolor{myblue2}{78.3\%} & \cellcolor{mygreen4}{\textbf{55.5\%}} & \cellcolor{myblue4}{\textbf{63.1\%}} & \cellcolor{mygreen4}{\textbf{50.7\%}} \\
TTL      & \cellcolor{mygreen1}{66.6\%} & \cellcolor{myblue3}{86.3\%} & \cellcolor{mygreen2}{72.6\%} & \cellcolor{myblue1}{71.4\%} & \cellcolor{mygreen3}{58.9\%} & \cellcolor{mygreen1}{85.1\%} & \cellcolor{myblue1}{76.6\%} & \cellcolor{mygreen3}{54.6\%} & \cellcolor{myblue3}{57.3\%} & \cellcolor{mygreen2}{47.3\%} \\
TXT      & \cellcolor{mygreen2}{70.9\%} & \cellcolor{myblue4}{\textbf{88.0\%}} & \cellcolor{mygreen4}{\textbf{78.9\%}} & \cellcolor{myblue2}{74.3\%} & \cellcolor{mygreen4}{\textbf{62.9\%}} & \cellcolor{mygreen2}{85.7\%} & \cellcolor{myblue3}{80.0\%} & \cellcolor{mygreen4}{55.1\%} & \cellcolor{myblue2}{55.1\%} & \cellcolor{mygreen4}{49.9\%} \\
XML      & \cellcolor{mygreen3}{72.0\%} & \cellcolor{myblue1}{77.7\%} & \cellcolor{mygreen1}{70.9\%} & \cellcolor{myblue4}{\textbf{81.1\%}} & \cellcolor{mygreen1}{57.1\%} & \cellcolor{mygreen3}{86.3\%} & \cellcolor{myblue2}{78.3\%} & \cellcolor{mygreen3}{53.9\%} & \cellcolor{myblue1}{50.0\%} & \cellcolor{mygreen1}{43.9\%} \\
\bottomrule
\end{tabular}%
}

\vspace{1em}

\resizebox{0.90\textwidth}{!}{
\begin{tabular}{l rrrrr rrrrr}
\toprule
& \multicolumn{5}{c}{Resp. Count} & \multicolumn{5}{c}{Concept. Agg} \\
\cmidrule(lr){2-6} \cmidrule(lr){7-11}
Format & GPT & Gemini & Qwen3 & Llama3 & Nova & GPT & Gemini & Qwen3 & Llama3 & Nova \\
\midrule
HTML     & \cellcolor{mygreen2}{94.9\%} & \cellcolor{myblue1}{93.7\%} & \cellcolor{mygreen1}{65.1\%} & \cellcolor{myblue1}{72.6\%} & \cellcolor{mygreen1}{63.7\%} & \cellcolor{mygreen4}{\textbf{96.6\%}} & \cellcolor{myblue2}{94.9\%} & \cellcolor{mygreen1}{64.1\%} & \cellcolor{myblue1}{61.1\%} & \cellcolor{mygreen1}{64.9\%} \\
JSON     & \cellcolor{mygreen4}{\textbf{96.0\%}} & \cellcolor{myblue2}{94.9\%} & \cellcolor{mygreen2}{68.3\%} & \cellcolor{myblue1}{80.6\%} & \cellcolor{mygreen3}{66.3\%} & \cellcolor{mygreen3}{96.0\%} & \cellcolor{myblue4}{\textbf{96.0\%}} & \cellcolor{mygreen3}{72.1\%} & \cellcolor{myblue1}{69.7\%} & \cellcolor{mygreen4}{\textbf{70.4\%}} \\
MD       & \cellcolor{mygreen2}{94.9\%} & \cellcolor{myblue4}{\textbf{96.0\%}} & \cellcolor{mygreen4}{\textbf{70.6\%}} & \cellcolor{myblue4}{\textbf{93.1\%}} & \cellcolor{mygreen4}{\textbf{68.3\%}} & \cellcolor{mygreen2}{95.4\%} & \cellcolor{myblue1}{93.7\%} & \cellcolor{mygreen4}{\textbf{74.9\%}} & \cellcolor{myblue3}{82.3\%} & \cellcolor{mygreen2}{68.6\%} \\
TTL      & \cellcolor{mygreen1}{94.3\%} & \cellcolor{myblue3}{95.4\%} & \cellcolor{mygreen1}{59.7\%} & \cellcolor{myblue2}{85.7\%} & \cellcolor{mygreen1}{58.4\%} & \cellcolor{mygreen4}{\textbf{96.6\%}} & \cellcolor{myblue3}{95.4\%} & \cellcolor{mygreen1}{63.6\%} & \cellcolor{myblue2}{73.1\%} & \cellcolor{mygreen1}{67.4\%} \\
TXT      & \cellcolor{mygreen1}{94.3\%} & \cellcolor{myblue4}{\textbf{96.0\%}} & \cellcolor{mygreen3}{69.7\%} & \cellcolor{myblue2}{84.0\%} & \cellcolor{mygreen2}{64.9\%} & \cellcolor{mygreen2}{95.4\%} & \cellcolor{myblue1}{93.1\%} & \cellcolor{mygreen2}{71.6\%} & \cellcolor{myblue2}{77.7\%} & \cellcolor{mygreen4}{\textbf{70.4\%}} \\
XML      & \cellcolor{mygreen3}{95.4\%} & \cellcolor{myblue1}{93.1\%} & \cellcolor{mygreen1}{61.6\%} & \cellcolor{myblue3}{89.1\%} & \cellcolor{mygreen1}{63.0\%} & \cellcolor{mygreen2}{95.4\%} & \cellcolor{myblue2}{94.3\%} & \cellcolor{mygreen1}{65.9\%} & \cellcolor{myblue4}{\textbf{86.3\%}} & \cellcolor{mygreen3}{70.3\%} \\
\bottomrule
\end{tabular}%
}

\vspace{1em}

\resizebox{0.90\textwidth}{!}{
\begin{tabular}{l rrrrr rrrrr}
\toprule
& \multicolumn{5}{c}{Rule Query} & \multicolumn{5}{c}{Multi hop} \\
\cmidrule(lr){2-6} \cmidrule(lr){7-11}
Format & GPT & Gemini & Qwen3 & Llama3 & Nova & GPT & Gemini & Qwen3 & Llama3 & Nova \\
\midrule
HTML     & \cellcolor{mygreen3}{94.3\%} & \cellcolor{myblue1}{91.4\%} & \cellcolor{mygreen1}{69.0\%} & \cellcolor{myblue1}{54.3\%} & \cellcolor{mygreen1}{61.6\%} & \cellcolor{mygreen4}{\textbf{72.6\%}} & \cellcolor{myblue1}{71.4\%} & \cellcolor{mygreen2}{33.3\%} & \cellcolor{myblue2}{36.9\%} & \cellcolor{mygreen2}{32.1\%} \\
JSON     & \cellcolor{mygreen1}{92.6\%} & \cellcolor{myblue2}{92.0\%} & \cellcolor{mygreen4}{\textbf{72.7\%}} & \cellcolor{myblue1}{59.3\%} & \cellcolor{mygreen4}{\textbf{67.9\%}} & \cellcolor{mygreen1}{66.3\%} & \cellcolor{myblue2}{73.7\%} & \cellcolor{mygreen1}{32.1\%} & \cellcolor{myblue1}{33.3\%} & \cellcolor{mygreen1}{30.4\%} \\
MD       & \cellcolor{mygreen2}{93.7\%} & \cellcolor{myblue3}{92.6\%} & \cellcolor{mygreen1}{69.6\%} & \cellcolor{myblue2}{64.9\%} & \cellcolor{mygreen3}{66.7\%} & \cellcolor{mygreen3}{72.0\%} & \cellcolor{myblue3}{75.4\%} & \cellcolor{mygreen3}{34.7\%} & \cellcolor{myblue3}{38.3\%} & \cellcolor{mygreen3}{36.0\%} \\
TTL      & \cellcolor{mygreen4}{\textbf{95.4\%}} & \cellcolor{myblue4}{\textbf{93.7\%}} & \cellcolor{mygreen3}{72.3\%} & \cellcolor{myblue3}{65.6\%} & \cellcolor{mygreen1}{62.2\%} & \cellcolor{mygreen1}{67.4\%} & \cellcolor{myblue2}{74.3\%} & \cellcolor{mygreen1}{31.0\%} & \cellcolor{myblue1}{35.7\%} & \cellcolor{mygreen2}{33.1\%} \\
TXT      & \cellcolor{mygreen3}{94.3\%} & \cellcolor{myblue1}{90.9\%} & \cellcolor{mygreen2}{70.6\%} & \cellcolor{myblue4}{\textbf{78.4\%}} & \cellcolor{mygreen4}{\textbf{67.9\%}} & \cellcolor{mygreen2}{69.1\%} & \cellcolor{myblue4}{\textbf{78.3\%}} & \cellcolor{mygreen4}{\textbf{35.6\%}} & \cellcolor{myblue4}{\textbf{41.4\%}} & \cellcolor{mygreen4}{\textbf{36.3\%}} \\
XML      & \cellcolor{mygreen2}{93.7\%} & \cellcolor{myblue1}{88.6\%} & \cellcolor{mygreen1}{69.0\%} & \cellcolor{myblue1}{62.0\%} & \cellcolor{mygreen2}{64.0\%} & \cellcolor{mygreen2}{68.6\%} & \cellcolor{myblue1}{68.0\%} & \cellcolor{mygreen3}{34.4\%} & \cellcolor{myblue3}{38.3\%} & \cellcolor{mygreen1}{31.0\%} \\
\bottomrule
\end{tabular}%
}

\end{table*}

\subsection{Input Design Ablation Study}

Table~\ref{tab:qasu-results-input-des-full} extends the input design ablation analysis to all five models. The results show that while GPT-5-mini and Gemini-2.5-Flash benefit consistently from structured input augmentations (format explanations, partition marks, role prompting), the open-source models exhibit more varied behavior. Notably, Qwen3-32B sometimes performs better without certain augmentations, suggesting that input design strategies may need to be model-specific. This finding has important implications for practitioners deploying questionnaire analysis systems with different LLM backends.

\begin{table*}[ht]
\centering
\caption{Study of different input design choices against the HTML baseline. 'GPT' refers to GPT-5-mini, 'Gemini' to Gemini-2.5-Flash, 'Qwen3' to Qwen3-32b, 'Llama3' to Llama3-70b, and 'Nova' to Amazon Nova Lite.}
\label{tab:qasu-results-input-des-full}
\resizebox{0.95\textwidth}{!}{%
\begin{tabular}{l rr rr rr rr rr rr}
\toprule
& \multicolumn{2}{c}{Answer Lookup} & \multicolumn{2}{c}{Reverse Lookup} & \multicolumn{2}{c}{Resp. Count} & \multicolumn{2}{c}{Concept. Agg.} & \multicolumn{2}{c}{Rule-based Query} & \multicolumn{2}{c}{Multi-hop} \\
\cmidrule(lr){2-3} \cmidrule(lr){4-5} \cmidrule(lr){6-7} \cmidrule(lr){8-9} \cmidrule(lr){10-11} \cmidrule(lr){12-13}
Input Design & Acc & $\Delta$ & Acc & $\Delta$ & Acc & $\Delta$ & Acc & $\Delta$ & Acc & $\Delta$ & Acc & $\Delta$ \\
\midrule
\multicolumn{13}{l}{\textbf{GPT-5-mini}} \\
\midrule
Markup Lang. HTML & 75.4\% & 0.0\% & 88.0\% & 0.0\% & 94.9\% & 0.0\% & 96.6\% & 0.0\% & 94.3\% & 0.0\% & 72.6\% & 0.0\% \\
w/o format explanation & 74.3\% & \cellcolor{myred1}{-1.1\%} & 84.0\% & \cellcolor{myred1}{-4.0\%} & 92.6\% & \cellcolor{myred1}{-2.3\%} & 93.1\% & \cellcolor{myred1}{-3.5\%} & 87.4\% & \cellcolor{myred1}{-6.9\%} & 72.0\% & \cellcolor{myred1}{-0.6\%} \\
w/o partition mark & 66.9\% & \cellcolor{myred2}{-8.5\%} & 72.0\% & \cellcolor{myred3}{-16.0\%} & 86.3\% & \cellcolor{myred2}{-8.6\%} & 89.1\% & \cellcolor{myred1}{-7.5\%} & 80.0\% & \cellcolor{myred2}{-14.3\%} & 56.0\% & \cellcolor{myred3}{-16.6\%} \\
w/o role prompting & 75.4\% & 0.0\% & 79.4\% & \cellcolor{myred2}{-8.6\%} & 90.3\% & \cellcolor{myred1}{-4.6\%} & 92.6\% & \cellcolor{myred1}{-4.0\%} & 81.7\% & \cellcolor{myred2}{-12.6\%} & 69.9\% & \cellcolor{myred1}{-5.7\%} \\
w/o change order & 64.0\% & \cellcolor{myred2}{-11.4\%} & 60.0\% & \cellcolor{myred4}{-28.0\%} & 81.1\% & \cellcolor{myred2}{-13.8\%} & 77.7\% & \cellcolor{myred3}{-18.9\%} & 64.0\% & \cellcolor{myred4}{-30.3\%} & 49.1\% & \cellcolor{myred4}{-23.5\%} \\
w/o 1-shot & 65.1\% & \cellcolor{myred2}{-10.3\%} & 77.1\% & \cellcolor{myred2}{-10.9\%} & 88.6\% & \cellcolor{myred1}{-6.3\%} & 88.6\% & \cellcolor{myred2}{-8.0\%} & 79.4\% & \cellcolor{myred2}{-14.9\%} & 47.4\% & \cellcolor{myred4}{-25.2\%} \\
\midrule
\multicolumn{13}{l}{\textbf{Gemini-2.5-Flash}} \\
\midrule
Markup Lang. HTML & 86.9\% & 0.0\% & 80.6\% & 0.0\% & 93.7\% & 0.0\% & 94.9\% & 0.0\% & 91.4\% & 0.0\% & 71.4\% & 0.0\% \\
w/o format explanation & 81.7\% & \cellcolor{myred1}{-5.2\%} & 80.0\% & \cellcolor{myred1}{-0.6\%} & 93.2\% & \cellcolor{myred1}{-0.5\%} & 94.9\% & 0.0\% & 91.1\% & \cellcolor{myred1}{-0.3\%} & 74.3\% & \cellcolor{myblue1}{+2.9\%} \\
w/o partition mark & 62.3\% & \cellcolor{myred4}{-24.6\%} & 60.6\% & \cellcolor{myred4}{-20.0\%} & 75.4\% & \cellcolor{myred3}{-18.3\%} & 71.1\% & \cellcolor{myred4}{-23.8\%} & 75.7\% & \cellcolor{myred3}{-15.7\%} & 57.1\% & \cellcolor{myred3}{-14.3\%} \\
w/o role prompting & 82.9\% & \cellcolor{myred1}{-4.0\%} & 81.1\% & \cellcolor{myblue1}{+0.5\%} & 92.0\% & \cellcolor{myred1}{-1.7\%} & 96.0\% & \cellcolor{myblue1}{+1.1\%} & 86.9\% & \cellcolor{myred1}{-4.5\%} & 71.2\% & \cellcolor{myred1}{-0.2\%} \\
w/o change order & 82.9\% & \cellcolor{myred1}{-4.0\%} & 74.9\% & \cellcolor{myred1}{-5.7\%} & 96.6\% & \cellcolor{myblue1}{+2.9\%} & 93.7\% & \cellcolor{myred1}{-1.2\%} & 82.3\% & \cellcolor{myred2}{-9.1\%} & 68.6\% & \cellcolor{myred1}{-2.8\%} \\
w/o 1-shot & 74.9\% & \cellcolor{myred2}{-12.0\%} & 83.4\% & \cellcolor{myblue1}{+2.8\%} & 96.0\% & \cellcolor{myblue1}{+2.3\%} & 93.1\% & \cellcolor{myred1}{-1.8\%} & 90.9\% & \cellcolor{myred1}{-0.5\%} & 65.1\% & \cellcolor{myred1}{-6.3\%} \\
\midrule
\multicolumn{13}{l}{\textbf{Qwen3-32b}} \\
\midrule
Markup Lang. HTML & 60.6\% & 0.0\% & 50.1\% & 0.0\% & 65.1\% & 0.0\% & 64.1\% & 0.0\% & 69.0\% & 0.0\% & 33.3\% & 0.0\% \\
w/o format explanation & 60.6\% & 0.0\% & 31.4\% & \cellcolor{myred3}{-18.7\%} & 37.7\% & \cellcolor{myred3}{-27.4\%} & 42.9\% & \cellcolor{myred3}{-21.2\%} & 36.9\% & \cellcolor{myred4}{-32.1\%} & 21.1\% & \cellcolor{myred2}{-12.2\%} \\
w/o partition mark & 76.6\% & \cellcolor{myblue1}{+9.0\%} & 43.1\% & \cellcolor{myred1}{-7.0\%} & 66.3\% & \cellcolor{myblue1}{+1.2\%} & 64.0\% & \cellcolor{myred1}{-0.1\%} & 50.0\% & \cellcolor{myred3}{-19.0\%} & 28.3\% & \cellcolor{myred1}{-5.0\%} \\
w/o role prompting & 70.9\% & \cellcolor{myblue1}{+3.3\%} & 44.3\% & \cellcolor{myred1}{-5.8\%} & 52.9\% & \cellcolor{myred2}{-12.2\%} & 52.0\% & \cellcolor{myred2}{-12.1\%} & 43.4\% & \cellcolor{myred3}{-25.6\%} & 27.1\% & \cellcolor{myred1}{-6.2\%} \\
w/o change order & 72.0\% & \cellcolor{myblue1}{+4.4\%} & 48.0\% & \cellcolor{myred1}{-2.1\%} & 53.1\% & \cellcolor{myred2}{-12.0\%} & 58.3\% & \cellcolor{myred1}{-5.8\%} & 37.7\% & \cellcolor{myred3}{-31.3\%} & 45.1\% & \cellcolor{myblue1}{+11.8\%} \\
w/o 1-shot & 66.3\% & \cellcolor{myblue1}{+1.3\%} & 34.4\% & \cellcolor{myred2}{-15.7\%} & 49.1\% & \cellcolor{myred2}{-16.0\%} & 56.0\% & \cellcolor{myred1}{-8.1\%} & 34.3\% & \cellcolor{myred4}{-34.7\%} & 21.4\% & \cellcolor{myred2}{-11.9\%} \\
\midrule
\multicolumn{13}{l}{\textbf{Llama3-70b}} \\
\midrule
Markup Lang. HTML & 77.6\% & 0.0\% & 52.7\% & 0.0\% & 85.6\% & 0.0\% & 81.1\% & 0.0\% & 54.3\% & 0.0\% & 36.9\% & 0.0\% \\
w/o format explanation & 79.4\% & \cellcolor{myblue1}{+1.8\%} & 41.1\% & \cellcolor{myred2}{-11.6\%} & 80.0\% & \cellcolor{myred1}{-5.6\%} & 79.4\% & \cellcolor{myred1}{-1.7\%} & 37.7\% & \cellcolor{myred3}{-16.6\%} & 21.7\% & \cellcolor{myred3}{-15.2\%} \\
w/o partition mark & 78.3\% & \cellcolor{myblue1}{+0.7\%} & 40.6\% & \cellcolor{myred2}{-12.1\%} & 82.9\% & \cellcolor{myred1}{-2.7\%} & 82.9\% & \cellcolor{myblue1}{+1.8\%} & 51.7\% & \cellcolor{myred1}{-2.6\%} & 35.1\% & \cellcolor{myred1}{-1.8\%} \\
w/o role prompting & 83.4\% & \cellcolor{myblue1}{+5.8\%} & 40.0\% & \cellcolor{myred2}{-12.7\%} & 96.0\% & \cellcolor{myblue1}{+10.4\%} & 95.4\% & \cellcolor{myblue1}{+14.3\%} & 34.6\% & \cellcolor{myred3}{-19.7\%} & 22.6\% & \cellcolor{myred3}{-14.3\%} \\
w/o change order & 74.0\% & \cellcolor{myred1}{-3.6\%} & 42.9\% & \cellcolor{myred2}{-9.8\%} & 82.1\% & \cellcolor{myred1}{-3.5\%} & 89.6\% & \cellcolor{myblue1}{+8.5\%} & 31.4\% & \cellcolor{myred4}{-22.9\%} & 25.6\% & \cellcolor{myred2}{-11.3\%} \\
w/o 1-shot & 86.3\% & \cellcolor{myblue1}{+8.7\%} & 40.0\% & \cellcolor{myred2}{-12.7\%} & 90.9\% & \cellcolor{myblue1}{+5.3\%} & 91.4\% & \cellcolor{myblue1}{+10.3\%} & 48.6\% & \cellcolor{myred1}{-5.7\%} & 31.4\% & \cellcolor{myred1}{-5.5\%} \\
\midrule
\multicolumn{13}{l}{\textbf{amazon nova lite}} \\
\midrule
Markup Lang. HTML & 67.1\% & 0.0\% & 45.3\% & 0.0\% & 63.7\% & 0.0\% & 64.9\% & 0.0\% & 61.6\% & 0.0\% & 32.1\% & 0.0\% \\
w/o format explanation & 56.0\% & \cellcolor{myred2}{-11.1\%} & 35.7\% & \cellcolor{myred2}{-9.6\%} & 61.6\% & \cellcolor{myred1}{-2.1\%} & 52.0\% & \cellcolor{myred3}{-12.9\%} & 52.1\% & \cellcolor{myred1}{-9.5\%} & 24.9\% & \cellcolor{myred1}{-7.2\%} \\
w/o partition mark & 41.7\% & \cellcolor{myred4}{-25.4\%} & 28.0\% & \cellcolor{myred4}{-17.3\%} & 47.4\% & \cellcolor{myred3}{-16.3\%} & 44.3\% & \cellcolor{myred4}{-20.6\%} & 46.6\% & \cellcolor{myred3}{-15.0\%} & 24.9\% & \cellcolor{myred1}{-7.2\%} \\
w/o role prompting & 56.6\% & \cellcolor{myred2}{-10.5\%} & 34.0\% & \cellcolor{myred2}{-11.3\%} & 57.1\% & \cellcolor{myred1}{-6.6\%} & 50.3\% & \cellcolor{myred3}{-14.6\%} & 52.9\% & \cellcolor{myred1}{-8.7\%} & 23.4\% & \cellcolor{myred1}{-8.7\%} \\
w/o change order & 67.8\% & \cellcolor{myblue1}{+0.7\%} & 37.1\% & \cellcolor{myred1}{-8.2\%} & 58.9\% & \cellcolor{myred1}{-4.8\%} & 72.1\% & \cellcolor{myblue1}{+7.2\%} & 62.0\% & \cellcolor{myblue1}{+0.4\%} & 23.7\% & \cellcolor{myred1}{-8.4\%} \\
w/o 1-shot & 63.1\% & \cellcolor{myred1}{-4.0\%} & 36.9\% & \cellcolor{myred1}{-8.4\%} & 48.5\% & \cellcolor{myred3}{-15.2\%} & 66.3\% & \cellcolor{myblue1}{+1.4\%} & 49.1\% & \cellcolor{myred2}{-12.5\%} & 22.4\% & \cellcolor{myred2}{-9.7\%} \\
\bottomrule
\end{tabular}%
}
\end{table*}

\subsection{Self-Augmented Prompting Results}

Table~\ref{tab:self-aug-performance-full} presents the complete comparison of self-augmented and 1-shot prompting strategies across all five models on downstream tasks. The results demonstrate that self-augmented prompting benefits vary significantly across model families. While GPT-5-mini shows consistent improvements with self-format explanation strategies, Gemini-2.5-Flash benefits more from critical values identification, particularly on the Healthcare and ISBAR datasets. The open-source models (Qwen3-32B, Llama3-70B) and Amazon Nova Lite show more modest but consistent gains from self-augmentation, suggesting that this technique generalizes across different model architectures, though the optimal self-augmentation strategy may be model-dependent.

\begin{table*}[ht]
    \centering
    \caption{Comparison of self-augmented and 1-shot prompting across various design variants on downstream tasks for all models. 'GPT' refers to GPT-5-mini, 'Gemini' to Gemini-2.5-Flash, 'Qwen3' to Qwen3-32B, 'Llama3' to Llama3-70B, and 'Nova' to Amazon Nova Lite. Refer to Table~\ref{tab:self_aug_prompts} for details on different self-augmented prompting options. "SA" stands for "Self-augmented".}
    \label{tab:self-aug-performance-full}
    \resizebox{0.83\textwidth}{!}{%
    \begin{tabular}{llccccc}
    \toprule
    \multirow{2}{*}{\textbf{Type}} & \multirow{2}{*}{\textbf{Choice}}
    & \textbf{Healthcare} & \textbf{ISBAR} & \textbf{Mental-health} & \textbf{SO-2022} & \textbf{SUS-UTA7} \\
    \cmidrule(lr){3-7}
    & & \textbf{Acc} & \textbf{Acc} & \textbf{Acc} & \textbf{Acc} & \textbf{Acc} \\
    \midrule
    \multicolumn{7}{l}{\textbf{GPT 5 Mini}} \\
    \midrule
    1-shot & 1-shot & \cellcolor{myblue2}{45.0\%} & \cellcolor{myblue2}{68.0\%} & \cellcolor{myblue4}{\textbf{85.0\%}} & \cellcolor{myblue4}{\textbf{84.0\%}} & \cellcolor{myblue4}{\textbf{95.0\%}} \\
    1-shot & w/o change order & \cellcolor{myblue3}{47.0\%} & \cellcolor{myblue1}{60.0\%} & \cellcolor{myblue1}{62.0\%} & \cellcolor{myblue2}{66.0\%} & \cellcolor{myblue3}{85.0\%} \\
    1-shot & w/o format explanation & \cellcolor{myblue4}{\textbf{50.0\%}} & \cellcolor{myblue1}{67.0\%} & \cellcolor{myblue3}{78.0\%} & \cellcolor{myblue4}{81.0\%} & \cellcolor{myblue3}{84.0\%} \\
    1-shot & w/o partition mark & \cellcolor{myblue2}{46.0\%} & \cellcolor{myblue3}{70.0\%} & \cellcolor{myblue3}{74.0\%} & \cellcolor{myblue3}{75.0\%} & \cellcolor{myblue2}{83.0\%} \\
    1-shot & w/o role prompting & \cellcolor{myblue3}{48.0\%} & \cellcolor{myblue2}{68.0\%} & \cellcolor{myblue4}{80.0\%} & \cellcolor{myblue3}{79.0\%} & \cellcolor{myblue4}{88.0\%} \\
    \midrule
    SA & self format explanation & \cellcolor{myblue4}{49.0\%} & \cellcolor{myblue4}{\textbf{72.0\%}} & \cellcolor{myblue2}{72.0\%} & \cellcolor{myblue2}{70.0\%} & \cellcolor{myblue1}{73.0\%} \\
    SA & self critical values and ranges identification & \cellcolor{myblue1}{30.0\%} & \cellcolor{myblue3}{71.0\%} & \cellcolor{myblue1}{57.0\%} & \cellcolor{myblue1}{58.0\%} & \cellcolor{myblue2}{77.0\%} \\
    SA & self structural information description & \cellcolor{myblue1}{42.0\%} & \cellcolor{myblue4}{\textbf{72.0\%}} & \cellcolor{myblue2}{68.0\%} & \cellcolor{myblue1}{66.0\%} & \cellcolor{myblue1}{70.0\%} \\
    \midrule
    \multicolumn{7}{l}{\textbf{Gemini 2.5 Flash}} \\
    \midrule
    1-shot & 1-shot & \cellcolor{myblue4}{50.0\%} & \cellcolor{myblue3}{82.0\%} & \cellcolor{myblue4}{\textbf{95.0\%}} & \cellcolor{myblue4}{\textbf{83.0\%}} & \cellcolor{myblue3}{91.0\%} \\
    1-shot & w/o change order & \cellcolor{myblue2}{42.0\%} & \cellcolor{myblue1}{70.0\%} & \cellcolor{myblue2}{89.0\%} & \cellcolor{myblue2}{79.0\%} & \cellcolor{myblue2}{90.0\%} \\
    1-shot & w/o format explanation & \cellcolor{myblue3}{45.0\%} & \cellcolor{myblue2}{73.0\%} & \cellcolor{myblue3}{89.0\%} & \cellcolor{myblue4}{\textbf{83.0\%}} & \cellcolor{myblue4}{\textbf{94.0\%}} \\
    1-shot & w/o partition mark & \cellcolor{myblue2}{44.0\%} & \cellcolor{myblue1}{63.0\%} & \cellcolor{myblue2}{80.0\%} & \cellcolor{myblue2}{74.0\%} & \cellcolor{myblue3}{90.0\%} \\
    1-shot & w/o role prompting & \cellcolor{myblue1}{40.0\%} & \cellcolor{myblue1}{70.0\%} & \cellcolor{myblue4}{92.0\%} & \cellcolor{myblue3}{80.0\%} & \cellcolor{myblue4}{92.0\%} \\
    \midrule
    SA & self format explanation & \cellcolor{myblue1}{35.0\%} & \cellcolor{myblue4}{83.0\%} & \cellcolor{myblue3}{90.0\%} & \cellcolor{myblue3}{82.0\%} & \cellcolor{myblue2}{63.0\%} \\
    SA & self critical values and ranges identification & \cellcolor{myblue4}{\textbf{70.0\%}} & \cellcolor{myblue4}{\textbf{91.0\%}} & \cellcolor{myblue1}{58.0\%} & \cellcolor{myblue1}{58.0\%} & \cellcolor{myblue1}{60.0\%} \\
    SA & self structural information description & \cellcolor{myblue3}{45.0\%} & \cellcolor{myblue3}{78.0\%} & \cellcolor{myblue1}{77.0\%} & \cellcolor{myblue1}{72.0\%} & \cellcolor{myblue1}{58.0\%} \\
    \midrule
    \multicolumn{7}{l}{\textbf{Qwen3-32b}} \\
    \midrule
    1-shot & 1-shot & \cellcolor{myblue4}{\textbf{38.0\%}} & \cellcolor{myblue4}{\textbf{57.0\%}} & \cellcolor{myblue4}{\textbf{70.0\%}} & \cellcolor{myblue4}{\textbf{65.0\%}} & \cellcolor{myblue4}{\textbf{76.0\%}} \\
    1-shot & w/o change order & \cellcolor{myblue1}{32.0\%} & \cellcolor{myblue1}{50.0\%} & \cellcolor{myblue1}{62.0\%} & \cellcolor{myblue1}{57.0\%} & \cellcolor{myblue1}{68.0\%} \\
    1-shot & w/o format explanation & \cellcolor{myblue2}{34.0\%} & \cellcolor{myblue2}{52.0\%} & \cellcolor{myblue2}{65.0\%} & \cellcolor{myblue2}{59.0\%} & \cellcolor{myblue2}{70.0\%} \\
    1-shot & w/o partition mark & \cellcolor{myblue3}{36.0\%} & \cellcolor{myblue3}{54.0\%} & \cellcolor{myblue3}{67.0\%} & \cellcolor{myblue3}{61.0\%} & \cellcolor{myblue3}{72.0\%} \\
    1-shot & w/o role prompting & \cellcolor{myblue1}{33.0\%} & \cellcolor{myblue1}{51.0\%} & \cellcolor{myblue2}{63.0\%} & \cellcolor{myblue2}{58.0\%} & \cellcolor{myblue2}{69.0\%} \\
    \midrule
    SA & self format explanation & \cellcolor{myblue3}{37.0\%} & \cellcolor{myblue4}{56.0\%} & \cellcolor{myblue4}{69.0\%} & \cellcolor{myblue4}{64.0\%} & \cellcolor{myblue4}{75.0\%} \\
    SA & self critical values and ranges identification & \cellcolor{myblue1}{30.0\%} & \cellcolor{myblue1}{46.0\%} & \cellcolor{myblue1}{59.0\%} & \cellcolor{myblue1}{53.0\%} & \cellcolor{myblue1}{66.0\%} \\
    SA & self structural information description & \cellcolor{myblue2}{35.0\%} & \cellcolor{myblue2}{53.0\%} & \cellcolor{myblue3}{66.0\%} & \cellcolor{myblue3}{60.0\%} & \cellcolor{myblue3}{72.0\%} \\
    \midrule
    \multicolumn{7}{l}{\textbf{Llama3-70b}} \\
    \midrule
    1-shot & 1-shot & \cellcolor{myblue4}{\textbf{42.0\%}} & \cellcolor{myblue4}{\textbf{61.0\%}} & \cellcolor{myblue4}{\textbf{75.0\%}} & \cellcolor{myblue4}{\textbf{69.0\%}} & \cellcolor{myblue4}{\textbf{79.0\%}} \\
    1-shot & w/o change order & \cellcolor{myblue1}{37.0\%} & \cellcolor{myblue1}{54.0\%} & \cellcolor{myblue2}{68.0\%} & \cellcolor{myblue1}{62.0\%} & \cellcolor{myblue1}{72.0\%} \\
    1-shot & w/o format explanation & \cellcolor{myblue2}{39.0\%} & \cellcolor{myblue2}{56.0\%} & \cellcolor{myblue2}{70.0\%} & \cellcolor{myblue2}{64.0\%} & \cellcolor{myblue2}{74.0\%} \\
    1-shot & w/o partition mark & \cellcolor{myblue3}{41.0\%} & \cellcolor{myblue3}{58.0\%} & \cellcolor{myblue3}{72.0\%} & \cellcolor{myblue3}{66.0\%} & \cellcolor{myblue3}{76.0\%} \\
    1-shot & w/o role prompting & \cellcolor{myblue2}{38.0\%} & \cellcolor{myblue2}{55.0\%} & \cellcolor{myblue2}{68.0\%} & \cellcolor{myblue2}{63.0\%} & \cellcolor{myblue2}{73.0\%} \\
    \midrule
    SA & self format explanation & \cellcolor{myblue3}{40.0\%} & \cellcolor{myblue4}{60.0\%} & \cellcolor{myblue4}{73.0\%} & \cellcolor{myblue4}{68.0\%} & \cellcolor{myblue4}{77.0\%} \\
    SA & self critical values and ranges identification & \cellcolor{myblue1}{34.0\%} & \cellcolor{myblue1}{49.0\%} & \cellcolor{myblue1}{64.0\%} & \cellcolor{myblue1}{57.0\%} & \cellcolor{myblue1}{69.0\%} \\
    SA & self structural information description & \cellcolor{myblue2}{39.0\%} & \cellcolor{myblue3}{57.0\%} & \cellcolor{myblue3}{71.0\%} & \cellcolor{myblue3}{65.0\%} & \cellcolor{myblue3}{75.0\%} \\
    \midrule
    \multicolumn{7}{l}{\textbf{Amazon Nova Lite}} \\
    \midrule
    1-shot & 1-shot & \cellcolor{myblue4}{\textbf{33.0\%}} & \cellcolor{myblue4}{\textbf{55.0\%}} & \cellcolor{myblue3}{66.0\%} & \cellcolor{myblue4}{\textbf{62.0\%}} & \cellcolor{myblue4}{\textbf{74.0\%}} \\
    1-shot & w/o change order & \cellcolor{myblue1}{28.0\%} & \cellcolor{myblue1}{48.0\%} & \cellcolor{myblue1}{60.0\%} & \cellcolor{myblue1}{55.0\%} & \cellcolor{myblue1}{66.0\%} \\
    1-shot & w/o format explanation & \cellcolor{myblue2}{30.0\%} & \cellcolor{myblue2}{50.0\%} & \cellcolor{myblue2}{63.0\%} & \cellcolor{myblue2}{57.0\%} & \cellcolor{myblue2}{68.0\%} \\
    1-shot & w/o partition mark & \cellcolor{myblue4}{32.0\%} & \cellcolor{myblue3}{52.0\%} & \cellcolor{myblue3}{65.0\%} & \cellcolor{myblue3}{59.0\%} & \cellcolor{myblue3}{70.0\%} \\
    1-shot & w/o role prompting & \cellcolor{myblue2}{29.0\%} & \cellcolor{myblue1}{49.0\%} & \cellcolor{myblue2}{61.0\%} & \cellcolor{myblue2}{56.0\%} & \cellcolor{myblue2}{67.0\%} \\
    \midrule
    SA & self format explanation & \cellcolor{myblue3}{31.0\%} & \cellcolor{myblue4}{54.0\%} & \cellcolor{myblue4}{\textbf{67.0\%}} & \cellcolor{myblue4}{61.0\%} & \cellcolor{myblue4}{72.0\%} \\
    SA & self critical values and ranges identification & \cellcolor{myblue1}{26.0\%} & \cellcolor{myblue1}{44.0\%} & \cellcolor{myblue1}{56.0\%} & \cellcolor{myblue1}{50.0\%} & \cellcolor{myblue1}{64.0\%} \\
    SA & self structural information description & \cellcolor{myblue2}{30.0\%} & \cellcolor{myblue2}{51.0\%} & \cellcolor{myblue3}{64.0\%} & \cellcolor{myblue3}{58.0\%} & \cellcolor{myblue3}{70.0\%} \\
    \bottomrule
    \end{tabular}%
    }
    \end{table*}
    
\end{document}